\documentclass{article}

\PassOptionsToPackage{numbers,sort&compress}{natbib}
\usepackage[final]{neurips_2021}

\usepackage[utf8]{inputenc} % allow utf-8 input
\usepackage[T1]{fontenc}    % use 8-bit T1 fonts
\usepackage{hyperref}       % hyperlinks
\usepackage{url}            % simple URL typesetting
\usepackage{booktabs}       % professional-quality tables
\usepackage{amsfonts}       % blackboard math symbols
\usepackage{nicefrac}       % compact symbols for 1/2, etc.
\usepackage{microtype}      % microtypography
\usepackage[dvipsnames]{xcolor}         % colors
\usepackage{adjustbox}
\usepackage{makecell}
\usepackage{multirow}
\usepackage{array}
\usepackage{amsmath}
\usepackage{amsthm}
\usepackage{stmaryrd}
\usepackage{tikz}
\usepackage{subcaption}  
\usepackage{wrapfig}
\usepackage[normalem]{ulem}
\usepackage{algorithm}
\usepackage{algpseudocode}

\newtheorem{theorem}{Theorem}[section]

\DeclareMathOperator*{\argmin}{arg\,min}

\newcommand{\R}{\mathbb{R}}

\newcommand{\toptitlebar}{
  \hrule height 0.5em
  \vskip 0.25in
  \vskip -\parskip%
}
\newcommand{\bottomtitlebar}{
  \vskip 0.29in
  \vskip -\parskip
  \hrule height 0.10em
  \vskip 0.09in%
}

\newcommand\blfootnote[1]{%
  \begingroup
  \renewcommand\thefootnote{}\footnote{#1}%
  \addtocounter{footnote}{-1}%
  \endgroup
}

\newtheorem{innercustomthm}{Theorem}
\newenvironment{customthm}[1]
  {\renewcommand\theinnercustomthm{#1}\innercustomthm}
  {\endinnercustomthm}

\title{CorticalFlow: A Diffeomorphic Mesh Deformation Module for Cortical Surface Reconstruction}

\author{%
  L\'eo Lebrat$^{\dag}$\\
  CSIRO, QUT\\
  \texttt{leb026@csiro.au} \\
  \And
  Rodrigo Santa Cruz$^{\dag}$\\
  CSIRO, QUT\\
  \texttt{fon022@csiro.au} \\
  \And Fr\'ed\'eric de Gournay\\
  IMT - UMR5219
  \\
  \texttt{degourna@insa-toulouse.fr}
  \And
  Darren Fu\\
  UQ
  \And
  Pierrick Bourgeat\\
  CSIRO
  \And
  Jurgen Fripp\\
  CSIRO
  \And 
  Clinton Fookes\\
  QUT
  \And
  Olivier Salvado\\
  CSIRO, Data61.
}

\begin{document}

\maketitle

\begin{abstract}
In this paper we introduce CorticalFlow, a new geometric deep-learning model that, given a 3-dimensional image, learns to deform a reference template towards a targeted object. 
To conserve the template mesh’s topological properties,
we train our model over a set of diffeomorphic transformations. This new implementation of a flow Ordinary Differential Equation (ODE) framework benefits from a small GPU memory footprint, allowing the generation of surfaces with several hundred thousand vertices. To reduce topological errors introduced by its discrete resolution, we derive numeric conditions which improve the manifoldness of the predicted triangle mesh. 
To exhibit the utility of CorticalFlow, we demonstrate its performance for the challenging task of brain cortical surface reconstruction.
In contrast to current state-of-the-art, CorticalFlow produces superior surfaces while reducing the computation time from nine and a half minutes to one second. More significantly, CorticalFlow enforces the generation of anatomically plausible surfaces; the absence of which has been a major impediment restricting the clinical relevance of such surface reconstruction methods.
\end{abstract}

\section{Introduction}
\blfootnote{$\dag$ Equal contribution}
\blfootnote{Check our project web-page \url{https://lebrat.github.io/CorticalFlow/}}
The field of 3D shape reconstruction using deep learning techniques has attracted much attention. Recently, a plethora of methods have been developed for problems such as single-view object reconstruction~\cite{fahim2021single,zubic2021effective,niemeyer2020differentiable}, surface generation~\citep{girdhar2016learning,wu2018learning}, and meshing noisy point clouds~\citep{Hanocka2020p2m,liu2020meshing}. 
At first, these methods solely aimed to retrieve surface meshes as geometrically close as possible to the target shape. However, recent applications require generating regular meshes with a known genus, such as physics simulation, 3D-printing, and clinical analysis of anatomical surfaces~\citep{rodriguez2008framework,schaer2008surface,fischl1999cortical}. 

In this direction, three approaches in the literature stand out: DeepCSR~\citep{cruz2021deepcsr}, Voxel2Mesh~\citep{wickramasinghe2020voxel2mesh}, and Neural Mesh Flow (NMF)~\citep{gupta_neurips20_nmf}. 
DeepCSR first predicts implicit surface functions and then employs an iso-surface extraction method along with a topology correction algorithm to obtain genus-zero surfaces without handles or holes.
Voxel2Mesh extends the vertex-wise template deformation approach of \citet{wang2018pixel2mesh} by optimizing several mesh-smoothing penalty functions.
%and applying a topology correction routine within its adaptive mesh unpooling module to reduce the number of faulty vertices.
In contrast, NMF builds an invertible mapping that enforces topology conservation upon the resolution of an Ordinary Differential Equation (ODE) through a sequence of residual blocks called Neural Ordinary Differential Equation (NODE)~\citep{chen2018neuralode}.
However, these methods come with several limitations. The topology correction algorithm employed by DeepCSR is computationally expensive and is blind towards the anatomical validity of its reconstructions, which can result in implausible corrections and mesh artifacts. 
On the other hand, Voxel2Mesh and NMF rely on time-demanding and vertex-dependent building blocks such as graph convolution that do not scale up well
% in memory 
as the number of vertices in the template mesh increases to accommodate complex shapes.
In addition, the dynamics learned by the NODE model in NMF can be very complex and may lead to a non-diffeomorphic mapping resulting in self-intersections in the reconstructed mesh.

This paper introduces CorticalFlow (CF), a new geometric deep learning model that smoothly deforms a template mesh towards complex shapes producing high-resolution regular meshes. First, a simple 3D convolution neural network predicts a dense 3D flow field from a volumetric image with a modest GPU memory footprint. 
Second, we formulate a tractable mathematical framework to compute diffeomorphic mapping for each vertex by solving a flow ODE. 
We derive sufficient and comprehensible conditions for meeting the diffeomorphic properties of these transformations.
Finally, a sequence of these diffeomorphic mappings is composed to produce accurate high-resolution genus-zero regular meshes.

To evaluate our approach and compare it to existing techniques for regular surface reconstruction, we consider the problem of brain cortical surface reconstruction, which is an essential step for the analysis of brain morphometry in neurodegenerative diseases~\citep{du2007different} and psychological disorders~\citep{rimol2012cortical}. 
Given a 3D MRI of the brain, the goal is to describe the inner and outer surfaces of the brain cortex, which are both homeomorphic to a sphere.
Cortical surface reconstruction is challenging given the complexity, high resolution, and regularity required for the predicted meshes. 
In our experiments, CorticalFlow is more accurate than state-of-the-art methods, providing an average reduction of $17.38\%$ in Chamfer distance across all cortical surfaces compared to DeepCSR (the second-best performing method in this criteria). In terms of surface regularity, it surpasses NMF or Voxel2Mesh with an average reduction of at least $32.58\%$ of self-intersecting faces while handling template meshes with many more vertices. It is also faster and more memory-efficient than all of these competitors.

\section{Related Works \label{sec:rw}} % 1 page

\subsection{Geometric deep learning for surface reconstruction}

Supervised surface reconstruction can be broadly categorized according to the 3D shape representation used to encode the prediction as either volumetric, implicit surfaces representation, novel geometric primitives, or geometric~\citep{fahim2021single}.

Volumetric methods predict shapes encoded as a 3D grid of voxels containing discretized surface representations such as occupancy~\citep{choy20163d} and level-sets~\citep{michalkiewicz2019implicit}. 
From this representation, surfaces are obtained using iso-surface extraction methods, such as marching cubes~\citep{lewiner2003efficient}. 
While 3D volumetric processing is amenable to a convolutional neural network, the memory requirements are often a limitation to attain high-resolution reconstructions (it grows cubically with the voxel-grid resolution). To overcome this issue, approaches based on octrees~\citep{Hane2017,tatarchenko2017octree,wang2018adaptive} have been proposed to increase the output resolution from a voxel-grid of $32^3$ to $256^3$. Unfortunately, these approaches sacrifice speed and necessitate the redefinition of standard network operations such as convolution, pooling, and unpooling for this hierarchical data structure. Furthermore, as presented in \citep{cruz2021deepcsr}, even at this level of resolution, the precision is too coarse to capture the highly curved regions of the cortical surfaces.

Implicit surface methods alleviate resolution limitations of the volumetric methods by directly predicting surface representations like occupancy \citep{Mescheder_2019_CVPR}, signed distance \citep{park2019deepsdf,Xu:NIPS19}, and 3D Gaussians~\citep{genova2019learning} for points with continuous coordinates. 
This formulation allows synthesizing grids at an arbitrary resolution during inference with an easily implementable local refinement procedure, while training is performed stochastically over a small subset of sampled points. 
Following this approach, \citet{cruz2021deepcsr} proposed DeepCSR, the first geometric deep learning model for cortical surface reconstruction. 
Its main limitation is the difficulty to control the topology and mesh quality of the reconstructed surfaces, which hampers atrophy estimation used for neurodegenerative disease diagnosis~\citep{rodriguez2008framework,schaer2008surface,fischl1999cortical}. 
As a result, DeepCSR resorts to a computationally expensive topology correction algorithm to produce a final cortical surface almost free of artifacts and with a spherical topology.

Methods based on geometric primitives build a surface representation to approximate complex shapes as the union of these primitives. In this category, we can highlight the works of \citet{niu2018im2struct} and \citet{groueix2018papier} which propose to approximate complex object shapes with a collection of ``cuboids'' or ``surface patches''. Recent works by~\cite{deng2020cvxnet,chen2020bsp} revisit the convex decomposition idea and propose to reconstruct complex object shapes by predicting collections of convex parts. The former predicts a set of localized convex polytopes formed by their hyperplane parameters and a translation vector, while the latter predicts a binary space partitioning tree to reconstruct the target shape. While very promising in terms of information compactness, these approaches are challenged to generate cortical surfaces due to their varied curvatures, which require a large number of convex parts to produce accurate results.

Finally, geometric methods comprise techniques that allow estimating a high-resolution mesh by transforming a known template mesh~ \citep{wang2018pixel2mesh,pan2018residual,smith19a,Pan_2019_ICCV}.
Following this approach, \citet{wang2018pixel2mesh} propose a graph-convolution network to predict vertex-wise deformations of a spherical mesh while dynamically increasing its resolution with a point pooling process.
\citet{wickramasinghe2020voxel2mesh} extended this model for the reconstruction of smooth anatomical surfaces such as the liver or hippocampus for different image modalities. 
Topological errors are reduced using three different penalty functions in the loss function.
Recently, \citet{gupta_neurips20_nmf} leverage NODE blocks \citep{chen2018neuralode} to parameterize regular deformations that allow conserving the two-manifoldness property of the input template.

Indeed, the "\textit{manifoldness}" measures as non-manifold edges or non-manifold vertices, and defined by~\citet{gupta_neurips20_nmf}, are conserved by a deformable model which is not generating new vertices since those properties are inherited from the template mesh (only the vertices' positions are affected).
However, the normal consistency (non-manifold faces) is only conserved by homeomorphisms which can at most flip globally the faces' normal orientation.

%Moreover, a homeomorphism of a 2-manifold is topology-preserving so that the genus, the number of connected components and self intersections remain unchanged. However, this does not mean that an homeomorphism applied to the vertices of a triangle mesh (while keeping the connectivity) does not create self intersection.

\subsection{Generation of diffeomorphic mappings}

\begin{wrapfigure}[22]{l}{0.41\textwidth}
  \begin{center}
    \vspace*{-.9cm}\includegraphics[width=0.39\textwidth]{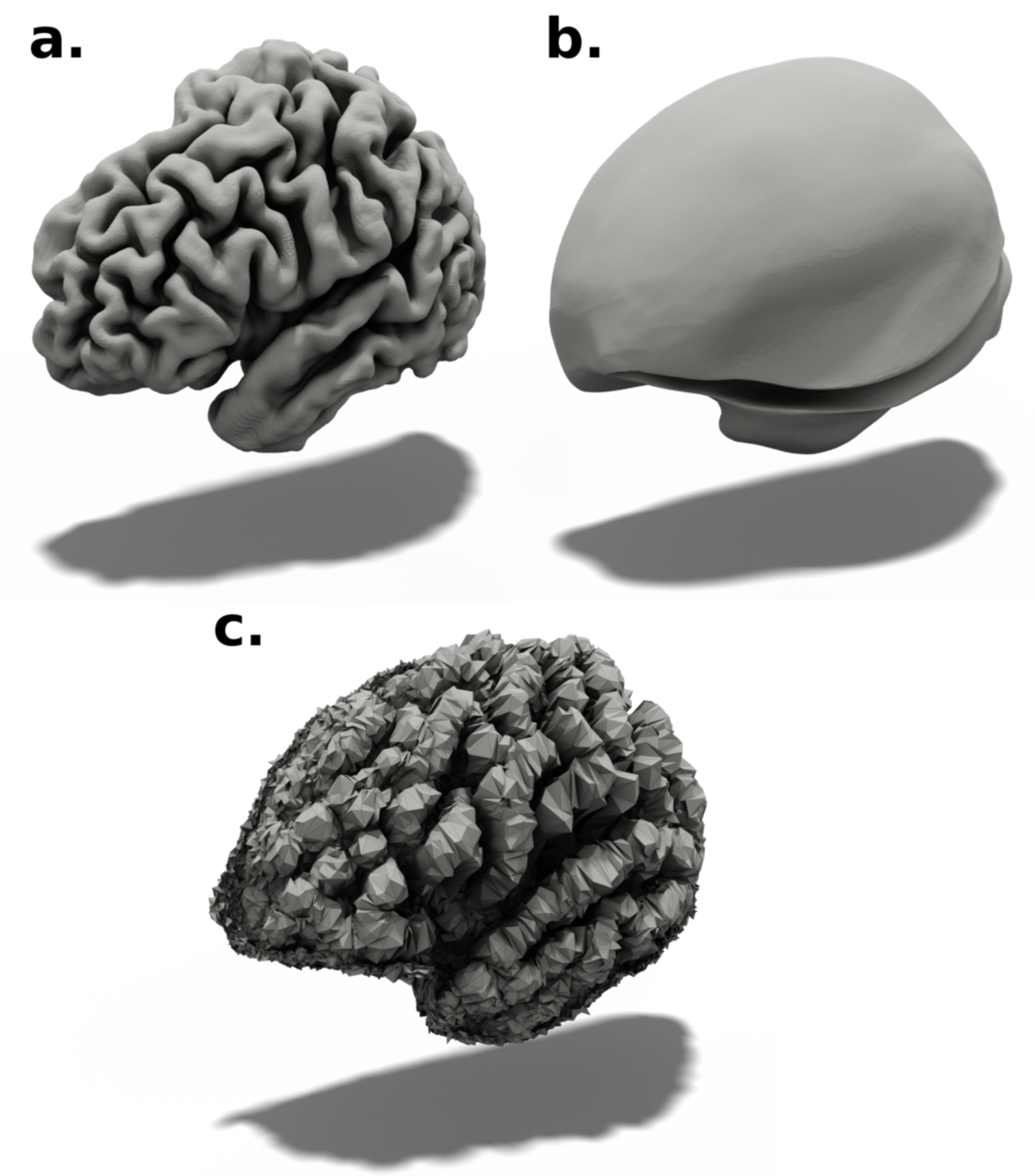}
  \end{center}
  \caption{Generating regular meshes of the left hemisphere of the brain cortical surface. \textbf{a.} \textit{CorticalFlow}, \textbf{b.} \textit{NMF} with a global feature descriptor \cite{gupta_neurips20_nmf}, \textbf{c.} \textit{UNet} without diffeomorphic parametrization or geometric penalization.}
  \label{fig:intro}
  \setcounter{figure}{1}
\end{wrapfigure}
The reconstruction of regular surfaces from a deformable model is a subtle trade off between finding the right parameterization or a suitable level of regularization during training. 
The delicacy of this problem is illustrated in Figure~1.
First, one can employ multiple penalty functions as Chamfer normal, normal consistency, Laplacian loss, or edge length loss~\cite{wang2018pixel2mesh,wickramasinghe2020voxel2mesh}. It is worth mentioning that without such penalizations, a deformable model will learn non-smooth deformations, which leads to irregular meshes (as shown in Figure~1.c). 
However, those penalizations simply encourage the reconstructed surface to be regular.
The second approach consists of parameterizing the set of learned deformations; this approach is favored in our paper since it allows stronger theoretical guarantees and is not subject to hyper-parameter tuning.
A natural framework to generate invertible deformations is to consider a flow ODE~\cite{ebin1970groups,trouve1995infinite}. This framework has been successfully applied in pattern recognition and image registration~\cite{dupuis1998variational,camion2001geodesic,arsigny2005polyrigid,ashburner2007fast}. The main idea is to consider the mapping as the solution at time $\tau$ of an initial value problem (IVP) of the form,

\begin{equation}\label{eqn:edo}
\frac{\text{d}\Phi(s;\mathbf{x})}{\text{d}s} = v\left(\Phi(s;\mathbf{x}),s\right), \text{ with } \Phi(0;\mathbf{x}) = \mathbf{x},
\end{equation}
under a regularity hypothesis on $v$ and upon boundedness of its support, and using the Picard-Lidel\"of theorem, one can show that a unique solution of this problem exists for $\tau \in \bar \R$.
In addition, the mapping $\mathbf{x} \mapsto \Phi(s;\mathbf{x})$ defines a family of diffeomorphisms~\cite{berger2012differential,arsigny2004processing} for all time $s \in [0,\tau]$ whose inverse can be computed through a backward integration.

When the vector field $v$ is constant over time \textit{i.e.} $v: \R^3 \rightarrow \R^3$, Equation~\eqref{eqn:edo} describes the Stationary Velocity Field (SVF) framework~\cite{arsigny2004processing,ashburner2007fast}. If $v$ is a time-varying vector field, the framework described in~\eqref{eqn:edo} becomes the LDDMM model~\cite{beg2005computing,vialard2012diffeomorphic,charon2013analysis,shen2019region}.

This generic framework has been successfully applied within deep learning methods for diffeomorphic image registration~\cite{dalca2019unsupervised,mok2020fast}, point-cloud completion~\cite{niemeyer2019occupancy}, single view reconstruction~\cite{gupta_neurips20_nmf} and to parameterize set of deformations~\cite{jiang2020shapeflow}.

The SVF formulation has been particularly fecund in medical deep-learning registration~\cite{dalca2019unsupervised,krebs2019learning,mok2020fast,zhang2020diffeomorphic}, where the resolution of Equation~\eqref{eqn:edo}, is performed using the scaling and squaring method~\cite{arsigny2005polyrigid,ashburner2007fast} to predict the displacement of each voxel-center and to compute the registered image.
However, this technique is not suitable for points that lie on non-regular coordinates. 
Naively, one can compute these mappings on a dense grid and then interpolate the deformation at non-regular coordinates. This simple approach is subject to two main limitations. Firstly, Equation~\eqref{eqn:edo} has to be solved in our context for millions of grid points where only a few hundred thousand vertices are displaced. Secondly, one cannot guarantee the invertibility of such a mapping with linear interpolation and one cannot compose provably several of those approximated mappings.

In~\cite{niemeyer2019occupancy,gupta_neurips20_nmf}, the problem is solved using a black-box neural ODE~\cite{chen2018neuralode} and by learning a neural vector field $v$. 
Despite allowing learning a time-dependent vector field, this approach has shown its limitations in our targeted application. 
Cortical surfaces are unique to each individual; indeed, the cortical folding patterns are similar to a fingerprint~\cite{mangin2004framework} and constitute a distinctive biometric for each individual.
Moreover, we observe that the classical approach, which consists of conditioning the neural ODE on a global feature descriptor of the input image, fails to provide satisfactory results for cortical surface reconstruction (see Figure~1.b. and the supplementary material). 
Instead, one has to equip each moving vertex with a local feature descriptor of the input image, limiting the number of vertices of the resulting mesh.

Our work lies at the intersection of these methods. We propose to extend the SVF framework for points lying in real-coordinates, with particular care given to the numerical affordability of the ODE solver. We define a multi-scale approach, so that the final deformation is the result of the composition of three successive deformations that allow to approach more complex mappings and alleviate the limitations of the one vector field SVF framework~\cite{lorenzi2013geodesics,feydy2020analyse}.
This framework is memory efficient, theoretically tractable, and can seamlessly handle large template meshes ($\approx 450\texttt{k}$ vertices).

%Moreover, in the context of the cortical surface, one has to blend these regularization terms nicely to be able to model with fidelity of the sulci and gyri since high regularization as normal consistency or Laplacian loss can prevent the model from reconstructing geometries with high curvature. 

\section{Method}

CorticalFlow (CF) is a multi-level deep learning architecture composed of several Diffeomorphic Mesh Deformation (DMD) modules. It takes as input a 3-dimensional Magnetic Resonance Image (MRI) of a patient brain denoted $\mathbf{I}$ (tensor of dimensions $H\times W \times D$) and a template $\mathcal{T}_i$ (where $i$ represents the degree of refinement of the template). CorticalFlow outputs the surface representation of an anatomical substructure by composing stackable diffeomorphic deformations generated by DMD modules. CorticalFlow with $k$ deformations ($\text{CF}^k$) can be written using the following recurrence,
\begin{align}\label{eqn:RecCF}
    \text{CF}^{1}_{\theta_1}(\mathbf{I},\mathcal{T}_1) &= \text{DMD}(\text{UNet}^{1}_{\theta_1}(\mathbf{I}),\mathcal{T}_1)) \nonumber \\
    \text{CF}^{i+1}_{\theta_{i+1}}(\mathbf{I},\mathcal{T}_{i+1}) &= \text{DMD}(\text{UNet}^{i+1}_{\theta_{i+1}}(\mathbf{U}_1^\frown \cdots \mathbf{U}_{i}^\frown\mathbf{I}),\text{CF}_{i}(\mathbf{I},\mathcal{T}_{i+1})) \quad \text{for 
    } i \geq 1, 
\end{align}
with $\mathbf{A}^{\frown}\mathbf{B}$ the channel-wise concatenation of the tensors $\mathbf{A}$ and $\mathbf{B}$ and where $\mathbf{U}_k$ denotes the output of the $k$-th $\text{UNet}^{k}_{\theta_k}$ parameterized by $\theta_k$.

In our paper we describe $\text{CF}^3$, a version of CorticalFlow with three stages where each stage is learned successively. CorticalFlow is trained in a supervised fashion, given a dataset $\mathcal{D}$ composed of pairs of MR-image $\mathbf{I}$ and triangle mesh $S$ representing a cortical structure and for $i \in \{ 1,2,3\}$ we optimize the following objective,
\begin{equation}\label{eqn:problem}
\argmin_{\theta_i} \sum_{(\textbf{I},\textbf{S})\in \mathcal{D}} \mathcal{L} \big(\text{CF}^{i}_{\theta_{i}}(\mathbf{I},\mathcal{T}_{i}),S).
\end{equation}

As training loss $\mathcal{L}(\cdot, \cdot)$, we minimize the mesh edge loss and Chamfer distance computed on point clouds of 150k points sampled from the predicted and ground-truth surfaces using random uniform sampling. The implementation of these losses and sampling algorithm are provided in the \texttt{PyTorch3D} library~\citep{ravi2020pytorch3d}. 

% CorticalFlow is a deep learning deformable model which predicts regular surfaces from a 3-dimensional image and a template mesh $\mathcal{T}$ for a specified downstream task.
% To enforce topological correctness, the set of deformations learned is parametrized to be diffeomorphic. The resulting learning problem can be stated as, 
% \begin{equation}\label{eqn:problem}
% \argmin_{\substack{ \theta \\ \Phi_\theta(\bullet) \in \text{Diffeo}(\Omega)}} \sum_{(\textbf{I},\textbf{S})\in \mathcal{D}} \mathcal{L} \big(\Phi_\theta(\textbf{I}) \circ \mathcal{T}, \textbf{S}\big),
% \end{equation}
% where $\Phi_\theta$ is a neural network that receives as input an image $\mathbf{I}$ and predicts a diffeomorphic transformation of the regular template surface mesh $\mathcal{T}$ towards the target surface mesh $\textbf{S}$.

% Our model comprises two fundamentals blocks, a flow vector field predictor and a Diffeomorphic Mesh Deformation module (DMD) that, given a flow vector field, outputs a diffeomorphic mapping.

\subsection{DMD Diffeomorphic Mesh Deformation module}
\label{sec:diffeomorphicMap}

The introduction of a Diffeomorphic Mesh Deformation module (DMD) is driven by the following classification of surfaces in $3$ dimensions:
\begin{theorem}
Suppose that $B$ is a smooth closed manifold of dimension $2$ embedded in $\mathbb{R}^3$. Suppose that $\Phi :[0,\tau]\times \mathbb{R}^3 \rightarrow \mathbb{R}^3$ is a family of homeomorphisms (continuous map such that for each $t$, the mapping $x\mapsto \Phi(t,x)$ is bijective with continuous inverse) , with $\Phi(0;\mathbf{x})=\mathbf{x}$. Then, for each $t$, the homotopy classes of $B$ and $\Phi_{\sharp B}(t) = \left\{ \Phi(t,y), y \in B \right\}$ are the same.
\end{theorem}
This theorem means that if $B$ is a sphere, the surface $\Phi_{\sharp B}(t)$ is of genus $0$ with no self-intersection. 

\subsubsection*{Existence and uniqueness of a solution to the continuous problem}
The DMD generation of diffeomorphic mappping relies on the resolution of a continuous flow ODE. 
For that purpose, let $v : \mathbf{x} \in \Omega \mapsto v(\mathbf{x}) \in \R^3$ be a constant over time vector field supported on the MRI space $\Omega$ and obtained by tri-linear interpolation of a feature map $\mathbf{U} \in \R^{H\times W \times D \times 3}$. Suppose that $v=0$ on $\partial \Omega$. The image origin is denoted by $O = (o_1,o_2,o_3)$ and denote by $d_i$ the interpolation spacing in the $i$-th direction such that $\Omega = [o_1,o_1 + d_1(H-1)] \times [o_2,o_2 + d_2(W-1)] \times [o_3,o_3 + d_3(D-1)]$.

\begin{theorem}
\label{thm:EDO}
Existence and uniqueness of the solution. Define $\Phi$ through the autonomous ODE,
\begin{equation}
\frac{\text{d}\Phi(s;\mathbf{x})}{\text{d}s} = v\left(\Phi(s;\mathbf{x})\right), \text{ with } \Phi(0;\mathbf{x}) = \mathbf{x}.
\end{equation}
Then $\Phi$ is uniquely defined on $\mathbb{R}\times \Omega$, is Lipschitz and for each $t$, the mapping $x\mapsto \Phi(t,x)$ is bijective with Lipschitz inverse. The proof of this result can be found in the supplementary material.
\end{theorem}

Being Lipschitz is more difficult to achieve than being merely continuous. Less formally, Theorem~\ref{thm:EDO} ensures that if $B$ is smooth, then for all $t$, the surface $\Phi_{\sharp B}(t)$ may, in the worst case,  have kinks. If $\Phi$ is only continuous and not Lipschitz, then the surface $\Phi_{\sharp B}(t)$ might have cusps that are more irregular than kinks. 
Note as well that the choice of the interpolation technique used to generate $v$ is pivotal since it drives the regularity of the right-hand side of Equation~\eqref{eqn:autoODE}. Indeed, $v$ Lipschitz's constant boundedness allows the definition of a solution to the continuous problem. More importantly, and as described in the next section, it rules the step-size to use for obtaining a stable numeric method.

\subsubsection*{Numerical resolution of ODE}
The DMD module solves for each vertices' position a flow ODE defined in Equation~\eqref{eqn:autoODE}, defined by $\Psi$ the numeric approximation of $\Phi$ by an explicit forward method, the invertibility of this discretisation is given by the following theorem
\begin{theorem}
\label{Discrete}
Let $v$ be $L$-Lipschitz. Define $\Psi$ as the Forward Euler approximation,
\begin{equation}\label{eqn:forwardEuler}
\Psi(h,x)=x+hv(x).
\end{equation}
$\Psi$ is a Lipschitz homeomorphism for each $h <  L^{-1}$.
\end{theorem}
As a result, in combination with Theorem~\ref{thm:EDO}, the surface $\Psi_{\sharp B}(h)$ is smooth as long as $hL< 1$. We note that the stability condition $hL<1$ is commonly used in Computational Fluid Dynamics~\cite{koumoutsakos2008flow}.
Less formally, this estimate tells us that the less regular (high gradient) $v$ is, the smaller the integration step should be chosen.
\begin{proof}
Let $h< L^{-1}$ and denote $f:x\mapsto x+hv(x)$. Then $f$ is a Lipschitz mapping and it is sufficient to prove the bijectivity of $f$. The injectivity stems from, for all $x\ne y$,
$$ \Vert f(x)-f(y)\Vert \ge  \Vert x-y\Vert - h \Vert v(x)-v(y)\Vert  \ge  \Vert x-y\Vert (1-hL) >0.$$
The surjectivity comes from the fact that for each $y$, the mapping $g:x\mapsto y-hv(x)$ is $hL$-Lipschitz with $hL<1$, hence is a contraction. By a fixed point theorem, $g$ admits a unique $x^\star$ solution to $g(x^\star)=x^\star$, or equivalently $y=f(x^\star)$. Hence, $f$ is surjective. 
\end{proof} 

\begin{wrapfigure}{L}{0.55\textwidth}
    \begin{minipage}{0.55\textwidth}
    \begin{algorithm}[H]
    \caption{$\text{DMD}$ module pseudo-code}\label{alg:DMD}
    \begin{algorithmic}
    \State \\
    \hspace*{-.40cm} \textbf{Input:} $\mathbf{U}\in\R^{H\times W \times D \times 3}$ \Comment{Discrete flow}\\
    \hspace*{-.40cm} \textbf{Input:} $\mathcal{T}$ with vertices $(V_i)_{i \in 1..m}$ \Comment{Triangle mesh}\\
    \hspace*{-.40cm} \textbf{Input:} $n \in \mathbb{N}^*$ \Comment{Number of integration steps}\\
    \hspace*{-.40cm} \textbf{Output:} Updated $(V_i)_{i \in 1..m}$\\
    \hspace*{-.40cm} $h \gets \frac 1 n$
    \Ensure $h < \frac 1 L$
    \For{$i \in \llbracket 1, m\rrbracket $}
    \For{$j \in \llbracket 1, n\rrbracket $}
    \State $V_i \gets \Psi(h,V_i)$
    \EndFor
    \EndFor
    \end{algorithmic}
    \end{algorithm}
\end{minipage}
\end{wrapfigure}

Theorem~\ref{Discrete} ensures that $\Psi_{\sharp B}(h)$ is a non-intersecting manifold. Unfortunately, there exists another layer of numerical approximation that prevents $\Psi_{\sharp B}(h)$ to have the desired topological properties. Indeed, suppose that $B$ is a sphere, one never computes $\Psi_{\sharp B}(h)$, but triangulates a sphere with vertices $V_i$, and evaluates the image of the vertices using Algorithm~\ref{alg:DMD}. A new mesh is formed by using the image of the vertices and the original connectivity. Since the perturbed edges of the mesh are not the images of the original edges by the Forward Euler scheme, the resulting mesh may self-intersect (we refer interested readers to Section~2.3 of our supplementary material).
Notwithstanding this limitation, we use the rule of thumb $hL\le 1$, and check that for all considered examples, and we have $hL \le 1$.

% Theorem~\ref{EDO} is concerned with the solution to an ODE, whereas in practice, only an approximation of the ODE is available at hand. We focus our attention on the numerical approximation of~\eqref{eqn:autoODE} by a forward Euler method. Specifically, we built a sequence of $N$ approximations $(\mathbf{V}^i_k)_k, k \in 1, \cdots, N$ for each vertex $\mathbf{V}^i_0$ of the original mesh. The vertices are updated using the following iteration,

% \begin{equation}\label{eqn:discreteInte}
% \mathbf{V}^{i}_{k+1} = \mathbf{V}^{i}_{k} + h v(\mathbf{V}^{i}_{k}), \quad  \text{ with } h=\frac{\tau}{N}. 
% \end{equation}

% More precise integration methods, like high-order Runge-Kutta methods, are available. However these methods require the evaluation of additional temporary points and are only useful because they decrease the local error if the underlying velocity field is regular enough. Since $v$ is only Lipschitz by construction, higher-order methods are not necessary in our situation.

\subsection{Network architecture}

\begin{figure}[t!]
    \centering
    \includegraphics[width=\textwidth]{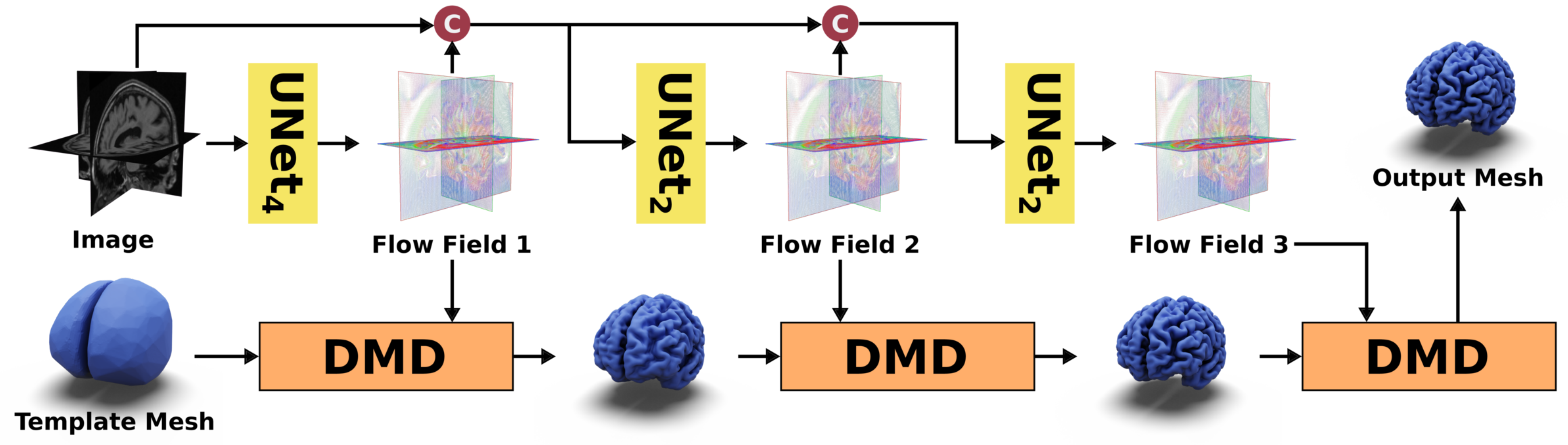}
    \caption{CorticalFlow architecture for the cortical surface reconstruction problem. Following the notations of Equation~\eqref{eqn:RecCF}, $\text{UNet}^{1}_{\theta_1}$ implements a $\textbf{UNet}_{\textbf{4}}$ layer and $\text{UNet}^{2,3}_{\theta_{2,3}}$ implements a $\textbf{UNet}_{\textbf{2}}$ layer. The architecture of $\textbf{UNet}_{\textbf{4}}$ and $\textbf{UNet}_{\textbf{2}}$ are described in the supplementary materials.}
   \label{fig:cortical_flow_arch}
\end{figure}

As shown in Figure~\ref{fig:cortical_flow_arch}, CorticalFlow consists of a chain of three deformations. Note that more deformation modules could be used, but we focus on three modules to have a fair comparison with existing techniques.
The first deformation module receives as input a volumetric image and outputs a flow vector field with the same dimensions using UNet-3D~\citep{ronneberger2015u}. This discrete flow vector field is integrated by the DMD module to compute smooth deformations as explained in Section~\ref{sec:diffeomorphicMap}.
The subsequent UNet-3D receives as input the image and the flow vector fields predicted by the previous deformation modules. The set of resulting mappings are composed to produce the final mesh.

For training CorticalFlow, we adopted a sequential approach where we train one deformation module at a time while freezing the weights of the others. 
We train the first deformation with a low-resolution template ($\mathcal{T}_1$) of $30\texttt{k}$ vertices and a UNet-3D architecture with four down/up-sampling levels. 
To train the second and third deformations, we increase the resolution of the template mesh to $135\texttt{k}$ and $435\texttt{k}$ vertices, for $\mathcal{T}_2$ and $\mathcal{T}_3$ respectively, and reduce the UNet-3D down/up-sampling levels to two. 
These networks are respectively labeled as $\textbf{UNet}_{\textbf{4}}$ and $\textbf{UNet}_{\textbf{2}}$ in Figure~\ref{fig:cortical_flow_arch} while their layer details are described in our supplementary material.
The choice of the architecture depth is motivated by the numerical conditions on the integration step-size derived in Section~\ref{sec:diffeomorphicMap}.

To verify the conditions of Theorem~\ref{Discrete}, it is essential to mention the use of template meshes with different resolutions and different UNet architectures. Indeed, the first block has to provide a large deformation resulting in a high $\|v\|$. To keep the Lipschitz constant $L$ small, one observes that the use of a low-resolution template with $30\texttt{k}$ vertices along with a deeper convolutional architecture forces the UNet to recover only coarse details and thus produces a flow vector field with a small $\nabla v$.
During the second and third deformation, more details and higher resolution folds can be learned, with templates composed of $135\texttt{k}$ and $435\texttt{k}$ vertices, respectively (see the ablation study presented in Section~1.1 of our supplementary material).
We empirically verify that this hierarchy of deformations was beneficial for producing a flow vector field $v$ with a small Lipschitz constant. This multi-step approach allows attaining up to $14.5$ times less self-intersection in comparison with Neural Mesh Flow (see Table~\ref{tab:benchmark}).

To generate the template mesh, we take the convex hull of all surfaces contained in the training dataset and remesh them uniformly using JIGSAW~\cite{engwirda2016off}. 
To achieve a different order of refinement, we use the midpoint subdivision algorithm implemented in MeshLab~\cite{LocalChapterEvents:ItalChap:ItalianChapConf2008:129-136}.
Model hyper-parameters and further implementation details are provided in the supplementary material.

\section{Experiments}

We benchmark CorticalFlow and other existing deep learning techniques on the cortical surface reconstruction problem. 
The goal is to estimate geometrically accurate and topologically correct triangular meshes for the inner and outer cortical surfaces from a given MRI. 
Like previous works \citep{Dahnke:NI2013:CAT,Dale:NI1999,Henschel:NI2020,Kim:NI2005:CLASP,Han2004:NI2004:CRUISE,cruz2021deepcsr}, these surfaces are further divided into the left and right brain hemispheres. 
See below a summary of the dataset, evaluated methods, and metrics used in our benchmark, in addition to the detailed discussion of the results summarized in Table~\ref{tab:benchmark}.

\begin{table}[ht!]
\centering
\resizebox{\textwidth}{!}{\begin{tabular}{l|cccccc|cccccc|}
\hline
& \multicolumn{6}{c|}{\large{Left Outer Surface}} & \multicolumn{6}{c|}{\large{Right Outer Surface }} \\
& CH($mm$) $\downarrow$ & HD($mm$) $\downarrow$ & CHN $\uparrow$ & \% SIF $\downarrow$ & DSC $\uparrow$ & VS $\uparrow$ & CH($mm$) $\downarrow$ & HD($mm$) $\downarrow$ & CHN $\uparrow$ & \% SIF $\downarrow$ & DSC $\uparrow$ & VS $\uparrow$\\
\cline{2-13}
 \makecell{\large{CorticalFlow} \\ 1.148 sec / 3.071 GB } & \makecell{$0.681$ \\ $(\pm 0.098)$} & \makecell{$0.802$ \\ $(\pm 0.049)$} & \makecell{$0.932$ \\ $(\pm 0.006)$} & \makecell{$0.686$ \\ $(\pm 0.469)$} & \makecell{$0.977$ \\ $(\pm 0.003)$} & \makecell{$0.993$ \\ $(\pm 0.006)$} & \makecell{$0.693$ \\ $(\pm 0.091)$} & \makecell{$0.815$ \\ $(\pm 0.046)$} & \makecell{$0.929$ \\ $(\pm 0.006)$} & \makecell{$1.239$ \\ $(\pm 0.629)$} & \makecell{$0.976$ \\ $(\pm 0.003)$} & \makecell{$0.994$ \\ $(\pm 0.005)$}\\
 \makecell{\large{DeepCSR} \\ 577.492 sec / 11.099 GB } & \makecell{$0.925$ \\ $(\pm 0.265)$} & \makecell{$0.898$ \\ $(\pm 0.135)$} & \makecell{$0.933$ \\ $(\pm 0.011)$} & \makecell{$0.000$ \\ $(\pm 0.000)$} & \makecell{$0.981$ \\ $(\pm 0.003)$} & \makecell{$0.993$ \\ $(\pm 0.005)$} & \makecell{$0.912$ \\ $(\pm 0.208)$} & \makecell{$0.895$ \\ $(\pm 0.125)$} & \makecell{$0.933$ \\ $(\pm 0.010)$} & \makecell{$0.000$ \\ $(\pm 0.000)$} & \makecell{$0.981$ \\ $(\pm 0.003)$} & \makecell{$0.993$ \\ $(\pm 0.005)$}\\
 \makecell{\large{NMF} \\ 42.808 sec / 14.431 GB } & \makecell{$1.557$ \\ $(\pm 0.200)$} & \makecell{$1.485$ \\ $(\pm 0.116)$} & \makecell{$0.885$ \\ $(\pm 0.008)$} & \makecell{$0.818$ \\ $(\pm 0.245)$} & \makecell{$0.953$ \\ $(\pm 0.005)$} & \makecell{$0.987$ \\ $(\pm 0.008)$} & \makecell{$1.772$ \\ $(\pm 0.177)$} & \makecell{$1.588$ \\ $(\pm 0.101)$} & \makecell{$0.871$ \\ $(\pm 0.008)$} & \makecell{$1.340$ \\ $(\pm 0.292)$} & \makecell{$0.946$ \\ $(\pm 0.005)$} & \makecell{$0.983$ \\ $(\pm 0.009)$}\\
 \makecell{\large{QuickNAT} \\ 13.003 sec / 9.627 GB } & \makecell{$3.732$ \\ $(\pm 1.971)$} & \makecell{$2.582$ \\ $(\pm 1.011)$} & \makecell{$0.874$ \\ $(\pm 0.024)$} & \makecell{$0.000$ \\ $(\pm 0.000)$} & \makecell{$0.956$ \\ $(\pm 0.006)$} & \makecell{$0.975$ \\ $(\pm 0.011)$} & \makecell{$4.282$ \\ $(\pm 2.293)$} & \makecell{$2.800$ \\ $(\pm 1.085)$} & \makecell{$0.870$ \\ $(\pm 0.024)$} & \makecell{$0.000$ \\ $(\pm 0.000)$} & \makecell{$0.954$ \\ $(\pm 0.006)$} & \makecell{$0.971$ \\ $(\pm 0.011)$}\\
 \makecell{\large{Voxel2Mesh} \\ 20.840 sec / 23.400 GB } & \makecell{$7.188$ \\ $(\pm 0.669)$} & \makecell{$3.893$ \\ $(\pm 0.218)$} & \makecell{$0.721$ \\ $(\pm 0.016)$} & \makecell{$0.857$ \\ $(\pm 0.260)$} & \makecell{$0.896$ \\ $(\pm 0.008)$} & \makecell{$0.976$ \\ $(\pm 0.011)$} & \makecell{$7.461$ \\ $(\pm 0.711)$} & \makecell{$3.955$ \\ $(\pm 0.227)$} & \makecell{$0.717$ \\ $(\pm 0.017)$} & \makecell{$0.833$ \\ $(\pm 0.260)$} & \makecell{$0.888$ \\ $(\pm 0.008)$} & \makecell{$0.967$ \\ $(\pm 0.010)$}\\
\hline
& \multicolumn{6}{c|}{\large{Left Inner Surface}} & \multicolumn{6}{c|}{\large{Right Inner Surface }} \\
& CH($mm$) $\downarrow$ & HD($mm$) $\downarrow$ & CHN $\uparrow$ & \% SIF $\downarrow$ & DSC $\uparrow$ & VS $\uparrow$ & CH($mm$) $\downarrow$ & HD($mm$) $\downarrow$ & CHN $\uparrow$ & \% SIF $\downarrow$ & DSC $\uparrow$ & VS $\uparrow$\\
\cline{2-13}
 \makecell{\large{CorticalFlow} \\ 1.148 sec / 3.071 GB } & \makecell{$0.608$ \\ $(\pm 0.098)$} & \makecell{$0.785$ \\ $(\pm 0.060)$} & \makecell{$0.941$ \\ $(\pm 0.007)$} & \makecell{$0.033$ \\ $(\pm 0.030)$} & \makecell{$0.962$ \\ $(\pm 0.005)$} & \makecell{$0.987$ \\ $(\pm 0.010)$} & \makecell{$0.599$ \\ $(\pm 0.093)$} & \makecell{$0.783$ \\ $(\pm 0.059)$} & \makecell{$0.942$ \\ $(\pm 0.007)$} & \makecell{$0.030$ \\ $(\pm 0.029)$} & \makecell{$0.962$ \\ $(\pm 0.005)$} & \makecell{$0.987$ \\ $(\pm 0.010)$}\\
 \makecell{\large{DeepCSR} \\ 577.492 sec / 11.099 GB } & \makecell{$0.653$ \\ $(\pm 0.138)$} & \makecell{$0.767$ \\ $(\pm 0.063)$} & \makecell{$0.956$ \\ $(\pm 0.006)$} & \makecell{$0.000$ \\ $(\pm 0.000)$} & \makecell{$0.963$ \\ $(\pm 0.006)$} & \makecell{$0.985$ \\ $(\pm 0.012)$} & \makecell{$0.634$ \\ $(\pm 0.139)$} & \makecell{$0.760$ \\ $(\pm 0.057)$} & \makecell{$0.956$ \\ $(\pm 0.006)$} & \makecell{$0.000$ \\ $(\pm 0.000)$} & \makecell{$0.964$ \\ $(\pm 0.006)$} & \makecell{$0.986$ \\ $(\pm 0.011)$}\\
 \makecell{\large{NMF} \\ 42.808 sec / 14.431 GB } & \makecell{$1.404$ \\ $(\pm 0.185)$} & \makecell{$1.447$ \\ $(\pm 0.140)$} & \makecell{$0.884$ \\ $(\pm 0.011)$} & \makecell{$0.355$ \\ $(\pm 0.168)$} & \makecell{$0.928$ \\ $(\pm 0.006)$} & \makecell{$0.984$ \\ $(\pm 0.012)$} & \makecell{$1.434$ \\ $(\pm 0.184)$} & \makecell{$1.463$ \\ $(\pm 0.135)$} & \makecell{$0.883$ \\ $(\pm 0.010)$} & \makecell{$0.436$ \\ $(\pm 0.188)$} & \makecell{$0.927$ \\ $(\pm 0.006)$} & \makecell{$0.984$ \\ $(\pm 0.012)$}\\
 \makecell{\large{QuickNAT} \\ 13.003 sec / 9.627 GB } & \makecell{$2.370$ \\ $(\pm 1.893)$} & \makecell{$1.836$ \\ $(\pm 0.734)$} & \makecell{$0.906$ \\ $(\pm 0.023)$} & \makecell{$0.000$ \\ $(\pm 0.000)$} & \makecell{$0.905$ \\ $(\pm 0.027)$} & \makecell{$0.924$ \\ $(\pm 0.031)$} & \makecell{$2.295$ \\ $(\pm 1.963)$} & \makecell{$1.788$ \\ $(\pm 0.725)$} & \makecell{$0.908$ \\ $(\pm 0.023)$} & \makecell{$0.000$ \\ $(\pm 0.000)$} & \makecell{$0.906$ \\ $(\pm 0.027)$} & \makecell{$0.931$ \\ $(\pm 0.031)$}\\
 \makecell{\large{Voxel2Mesh} \\ 20.840 sec / 23.400 GB } & \makecell{$5.968$ \\ $(\pm 0.657)$} & \makecell{$3.461$ \\ $(\pm 0.247)$} & \makecell{$0.703$ \\ $(\pm 0.017)$} & \makecell{$0.621$ \\ $(\pm 0.220)$} & \makecell{$0.850$ \\ $(\pm 0.009)$} & \makecell{$0.982$ \\ $(\pm 0.014)$} & \makecell{$6.172$ \\ $(\pm 0.719)$} & \makecell{$3.549$ \\ $(\pm 0.272)$} & \makecell{$0.707$ \\ $(\pm 0.017)$} & \makecell{$0.761$ \\ $(\pm 0.271)$} & \makecell{$0.842$ \\ $(\pm 0.010)$} & \makecell{$0.971$ \\ $(\pm 0.018)$}\\
\hline
\end{tabular}}
\caption{Cortical Surface Reconstruction Benchmark. Consider the evaluation metrics: chamfer distance (CH), Hausdorff distance (HD), chamfer normals (CHN), percentage of self-intersecting faces (\%SIF), Dice Score (DSC), and Volume Similarity (VS). $\downarrow$ indicates smaller metric value is better, while $\uparrow$ indicates greater metric value is better. We also report the inference runtime and GPU memory footprint required by the compared algorithms.}
\label{tab:benchmark}
\end{table}

\paragraph{Dataset.} We used the same MRIs, pseudo-ground-truth surfaces, and data splits as \citep{cruz2021deepcsr}. 
This dataset consists of 3,876 MRI images extracted from the Alzheimer’s Disease Neuroimaging Initiative (ADNI) \citep{Jack2008:ADNI} and their respective pseudo-ground-truth surfaces generated with the FreeSurfer V6.0 cross-sectional pipeline~\citep{fischl2002whole}. 
We train all methods on the training set ($\approx 60$\%) until their losses plateau on the validation set ($\approx 10$\%) and report their performance on the test set ($\approx 30$\%). 
We refer the reader to \citep{cruz2021deepcsr} and our supplementary material for full details on the dataset.

\paragraph{Evaluated Methods.} We compare CorticalFlow to the following methods: DeepCSR\footnote{DeepCSR official implementation retrieved from \url{https://bitbucket.csiro.au/projects/CRCPMAX/repos/deepcsr}}~\citep{cruz2021deepcsr}, Voxel2Mesh\footnote{Voxel2Mesh official implementation retrieved from \url{https://github.com/cvlab-epfl/voxel2mesh}}~\citep{wickramasinghe2020voxel2mesh}, NMF\footnote{NMF official implementation retrieved from \url{https://github.com/KunalMGupta/NeuralMeshFlow}}~\citep{gupta_neurips20_nmf}, and QuickNAT\footnote{QuickNAT official implementation retrieved from \url{https://github.com/ai-med/quickNAT_pytorch}}~\citep{roy2019quicknat}.  
As discussed in Section \ref{sec:rw}, DeepCSR is the state-of-the-art geometric deep learning model for cortical surface reconstruction, while Voxel2Mesh is a deformation-based model proposed to retrieve generic anatomical surfaces from volumetric medical images like MRIs and CT scans. 
Differently, NMF is a deformation-based model for single-view object reconstruction from 2D images. 
To adapt this model to our task whose input is a 3D MRI, we evaluate different 3D convolutional network backbones based on UNet \citep{ronneberger2015u}, ResNet \citep{he2016deep}, and Hypercolumn~\citep{cruz2021deepcsr}. 
The Hypercolumn backbone provides the best results thanks to its vertex-dependent features. As such, this is used as the NMF backbone in our benchmark. At the same time, the results for the other backbones are presented in our supplementary material. 
We also evaluate a baseline composed of a state-of-the-art brain segmentation model QuickNAT~\cite{roy2019quicknat} followed with the marching cubes to evaluate the surface. 
All of these methods were trained and evaluated using a NVIDIA P100 GPU and Intel Xeon (E5-2690) CPU, except Voxel2Mesh which required a NVIDIA RTX 3090 GPU due to its GPU memory requirements. % requested in  the neurips checklist

\paragraph{Evaluation Metrics.} We compare these methods for their geometric accuracy and surface regularity, as well as their time and space complexity. 
As a measure of geometric accuracy, we report the standard Chamfer distance (CH), Hausdorff distance (HD), and Chamfer normals (CHN). 
We compute these distances for point clouds of $200\texttt{k}$ points uniformly sampled from the predicted and target surfaces. 
As a measure of regularity, we compute the percentage of self-intersecting faces (\%SIF) using \texttt{PyMeshLab}~\cite{pymeshlab}.
We also report volumetric overlap metrics~\citep{taha2015metrics} including Dice Score (DSC) and Volume Similarity (VS) computed on the high-resolution rasterization ($4\times$the input MRI resolution) of the generated and ground-truth surfaces.

For the time and space complexity of the evaluated methods, we report their average inference time (in seconds) and inference GPU memory footprint (in GB) to reconstruct the four cortical surfaces, respectively.
% Refer to our supplementary materials for more details about these metrics.

\begin{figure}[ht!]
    \centering
    \includegraphics[width=\textwidth]{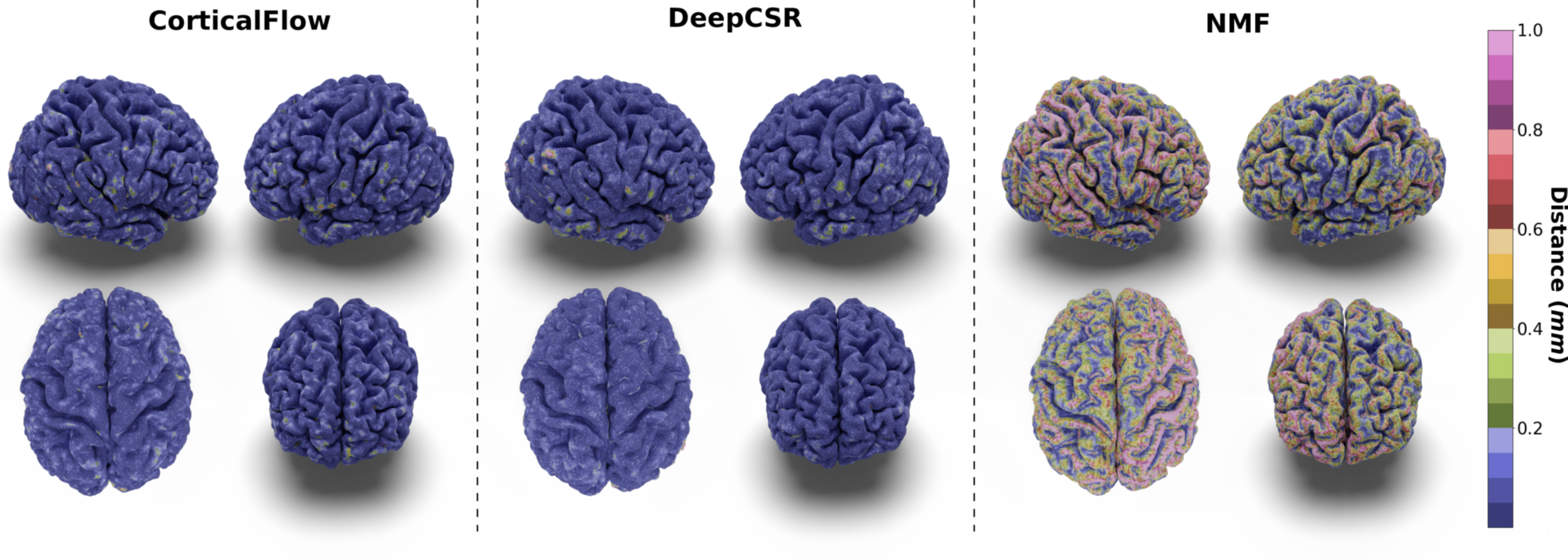}
    \caption{Predicted outer cortical surfaces color-coded with the distance to the pseudo-ground-truth surfaces. Here we present the results for the three best-performing methods in terms of geometric accuracy: CorticalFlow, DeepCSR, and NMF. See our supplementary materials for more examples.}
    \label{fig:surf_color_coded_ch}
\end{figure}

\paragraph{Results \& Discussion.} 
In our experiments, we noticed that CorticalFlow produces more geometrically accurate surfaces than the other methods. 
On average, it presents better geometric metrics across all the cortical surfaces. 
In addition, as shown in Figure~\ref{fig:surf_color_coded_ch}, CorticalFlow errors are smaller ($\leq 0.2mm$) and evenly spread across the surface compared to the other methods.
In contrast, NMF and DeepCSR can present substantial errors ($\geq 1mm$). The former has its error spread across the entire surface, while the latter can produce large errors at specific regions.

\begin{figure}[t!]
    \centering
    \includegraphics[width=\textwidth]{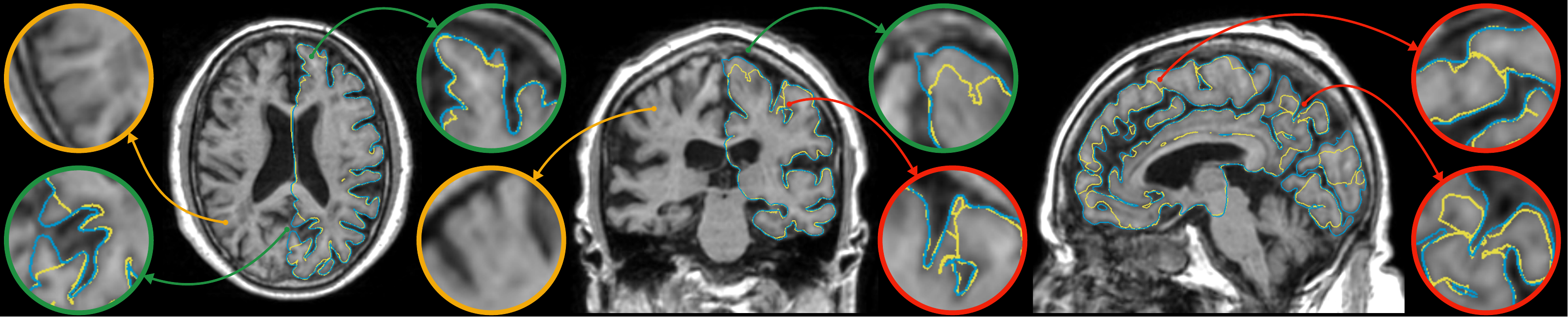}
    \caption{Slices of a blurry MRI and the outer surface delineation generated by \textcolor{Goldenrod}{FreeSurfer V6 (yellow contour)} and \textcolor{Cerulean}{CorticalFlow (blue contour)}. \textcolor{YellowOrange}{Orange circles highlight blurry MRI regions}, \textcolor{Green}{green circles highlight FreeSurfer's underestimated areas}, while \textcolor{red}{red circles highlight non-plausible predictions avoided by CorticalFlow thanks to the diffeomorphism of its predicted deformations}.}
    \label{fig:fs_error}
\end{figure}

CorticalFlow is also more robust than the competitors presenting lower error variation across individuals as suggested by the smaller standard deviation of the geometric metrics computed. 
Interestingly, CorticalFlow is also more robust to MRI artifacts even when the pseudo-ground-truth surface has poor quality. 
For instance, in Figure~\ref{fig:fs_error}, CorticalFlow predictions are still plausible for a blurry input MRI while FreeSurfer fails significantly to generate appropriate surfaces for the same input.
These examples support our claim that a regular parametrization allows us to reduce non-plausible and non-diffeomorphic predictions that our model cannot learn by construction.

\begin{figure}[h!]
    \centering
    \begin{subfigure}[t]{0.75\textwidth}
        \centering
        \includegraphics[width=\textwidth]{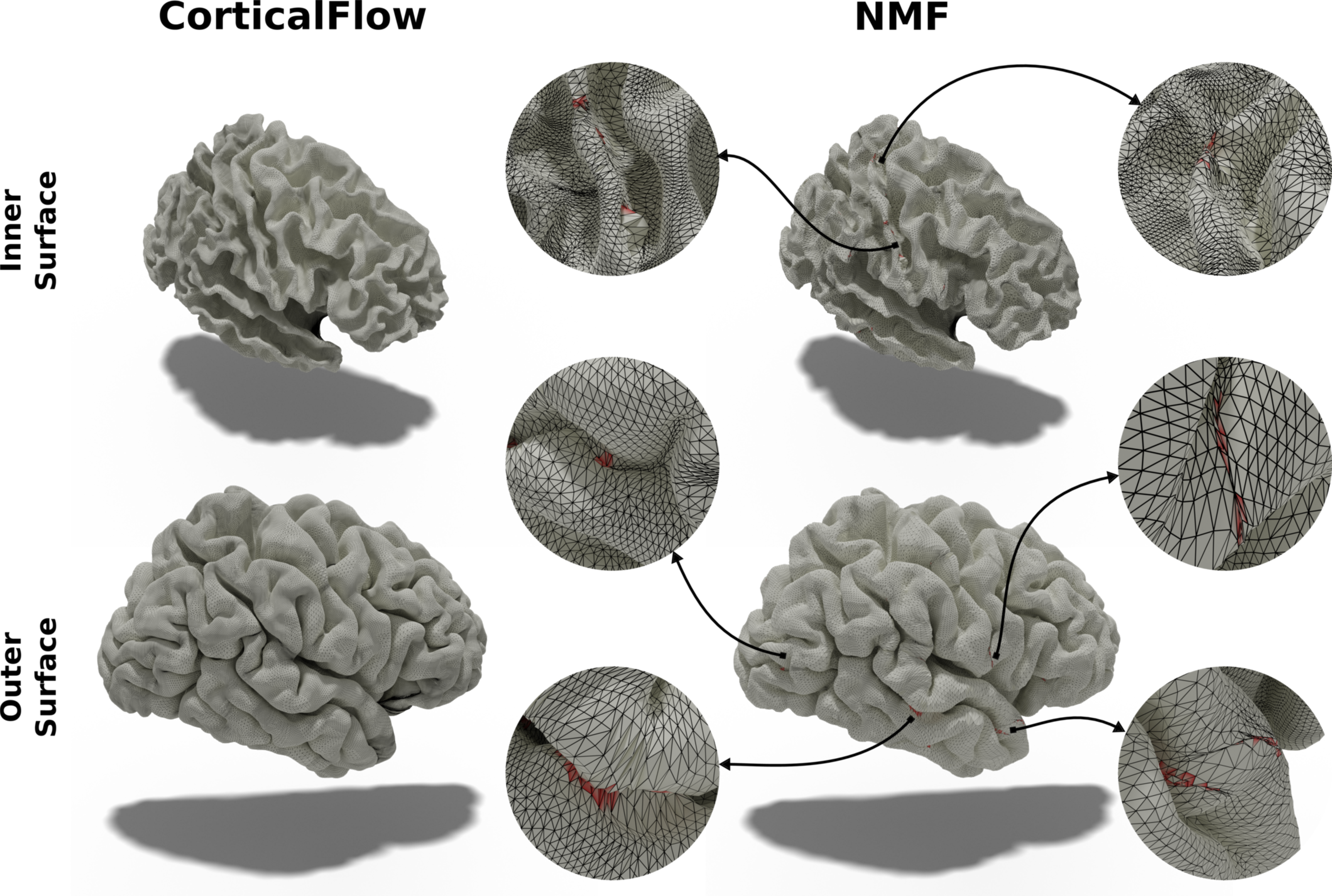}
        \caption{}
        \label{fig:surf_color_coded_si}
    \end{subfigure}%
    \hfill \vline \hfill
    \begin{subfigure}[t]{0.22\textwidth}
        \centering
        \includegraphics[width=\textwidth]{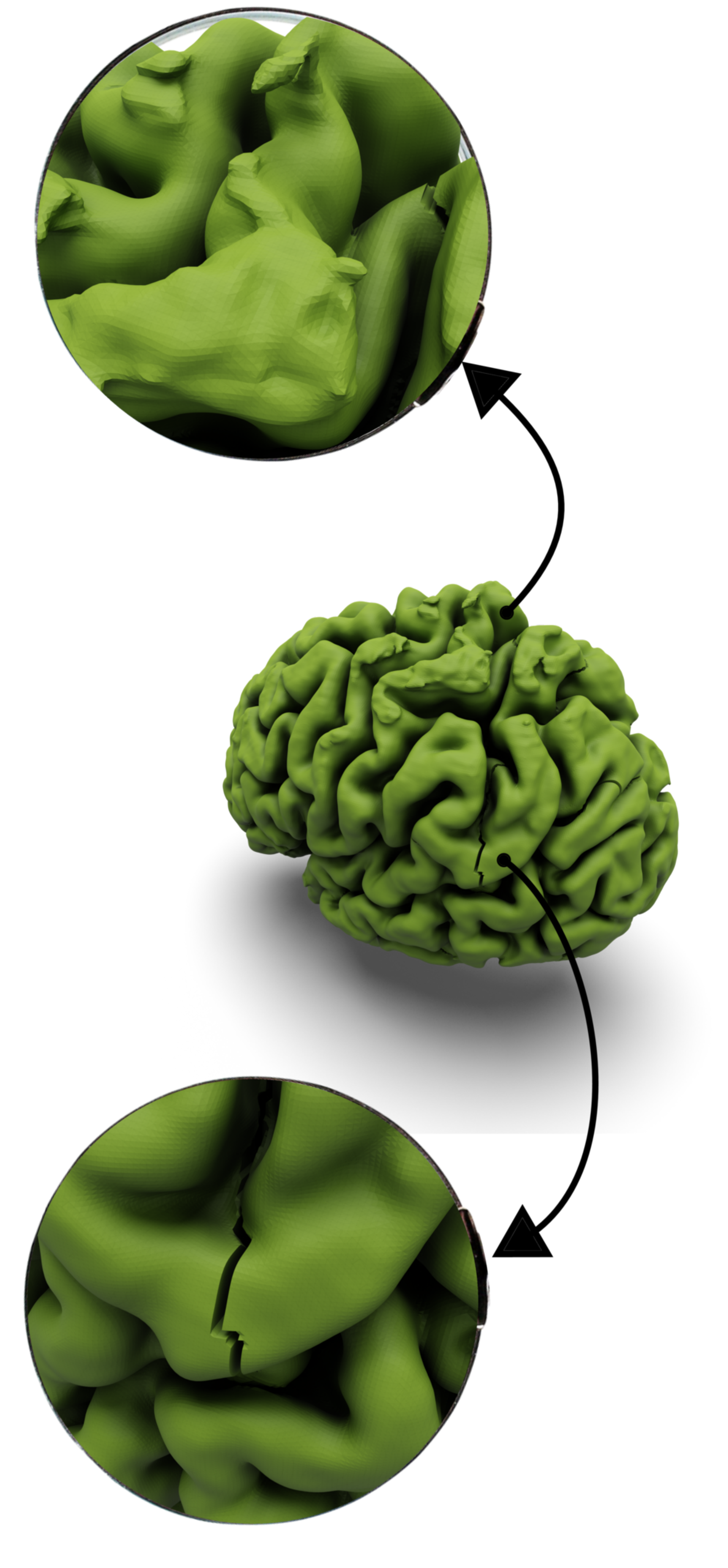}
        \caption{}
        \label{fig:deepcsr_topo_mistake}
    \end{subfigure}
    \caption{\subref{fig:surf_color_coded_si}: Predicted cortical surfaces by CorticalFlow and NMF with self-intersecting faces highlighted in red. \subref{fig:deepcsr_topo_mistake}: Significant mistakes generated by the topology correction algorithm used in the DeepCSR method.}
\end{figure}

CorticalFlow also generates triangular meshes with better properties than the evaluated methods. 
Compared to the deformation-based methods NMF and Voxel2Mesh, CorticalFlow predicted meshes are genus-zero surfaces and present a lower percentage of self-intersecting faces (mainly for the inner cortical surfaces). 
Figure~\ref{fig:surf_color_coded_si} presents examples of self-intersecting faces produced by CorticalFlow, which are contrasted with the NMF predicted mesh for the same input MRI.
The implicit-surface-based DeepCSR method does not produce a single self-intersecting face since it employs computationally expensive post-processing routines like topology correction~\citep{Bazin:2007} and iso-surface extraction. 
However, these post-processing routines do not take into account the input MRI which can generate non-plausible corrections on the output mesh as previously observed in \citet{Segonne:TMI07} and exemplified in Figure~\ref{fig:deepcsr_topo_mistake}. 
Similarly, the voxel-wise segmentation baseline (i.e., QuickNAT) is free of self-intersecting faces, but it does not produce genus-zero surfaces. 
Indeed, QuickNAT's predicted surfaces are composed of multiple connected components presenting many handles and holes which is not acceptable for the purpose of cortical surface reconstruction. Some examples of QuickNAT reconstructed cortical surfaces are presented in our supplementary material.
Therefore, we argue that CorticalFlow is the method of choice to reconstruct regular surfaces from volumetric images.

Due to its elemental construction (three UNet-3D backbones and an interpolation module for the integration), CorticalFlow remains highly efficient. 
It has a minimal GPU memory footprint and faster inference runtime while handling larger surfaces with more vertices both during training and inference.
This feature allows its deployment on low-end computers and embedded devices which is pivotal in many scenarios across public health and for commercialization of affordable AI healthcare solutions~\cite{cooley2020portable,paschali20193dq}.
 
\begin{figure}[h!]
    \centering
    \begin{subfigure}[t]{0.5\textwidth}
        \centering
        \includegraphics[width=0.88\textwidth]{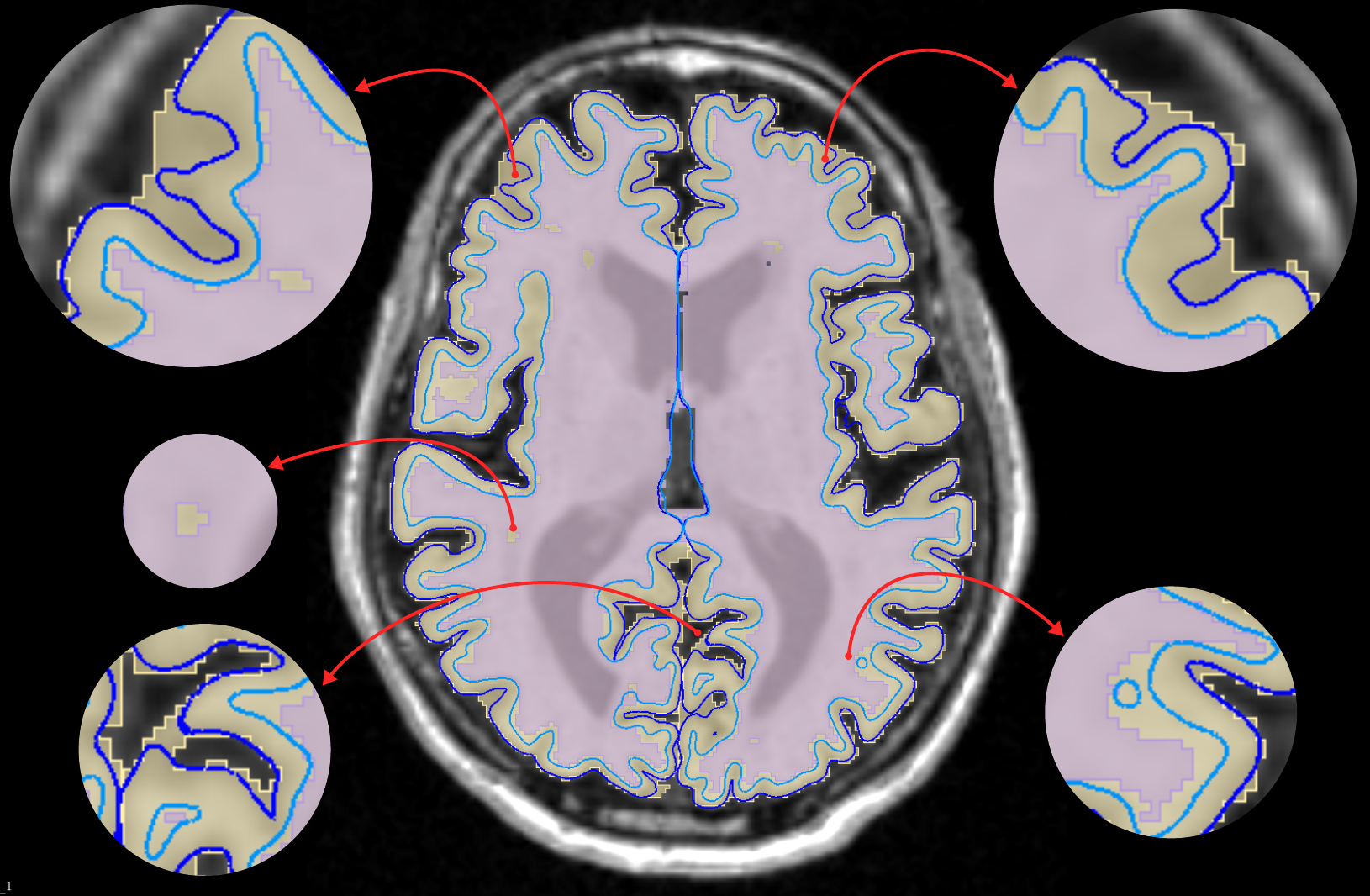}
        \caption{}
        \label{fig:app_highres_seg}
    \end{subfigure}%
    ~ 
    \begin{subfigure}[t]{0.5\textwidth}
        \centering
        \begin{tikzpicture}
        \node[anchor=south west,inner sep=0] at (0,0) {\includegraphics[width=\textwidth]{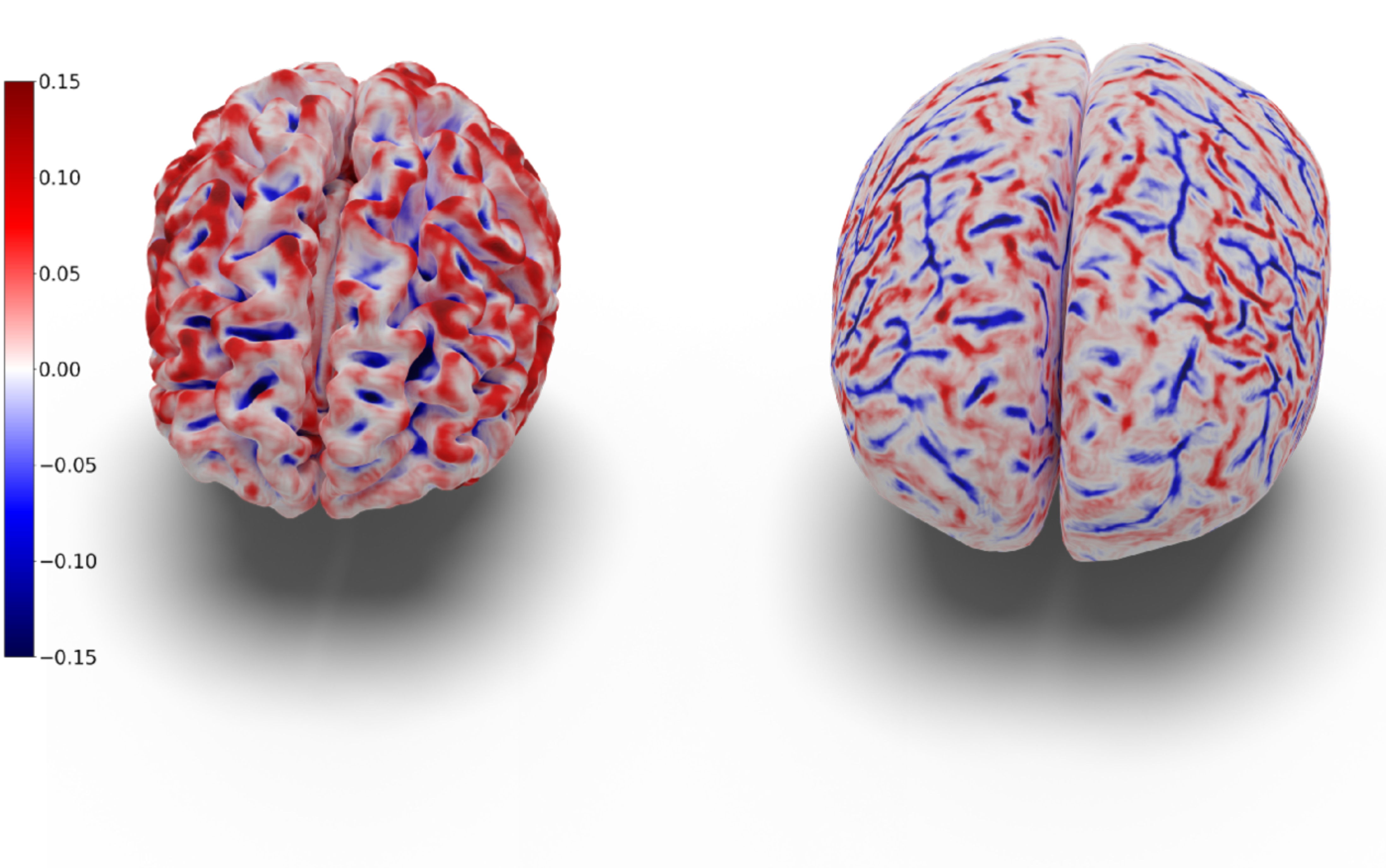}};
        \draw[ultra thick,->] (3.00,2.9) -- (4.1,2.9);
        \node[below] at (3.65,2.8) {\Large $\Phi^{-1}$};
        \end{tikzpicture}
        \caption{}
        \label{fig:app_curvature}
    \end{subfigure}
    \caption{\subref{fig:app_highres_seg}: QuickNAT segmentation (\textcolor{Lavender}{inner surface volume in pink} and  \textcolor{Goldenrod}{outer surface volume in yellow}) and CorticalFlow generated surfaces (\textcolor{CornflowerBlue}{inner surface in light blue} and  \textcolor{Blue}{outer surface in dark blue}). Note that the highlighted artifacts in QuickNAT segmentation can be easily surpassed by CorticalFlow thanks to its continuous representation of the region of interest. \subref{fig:app_curvature}: Outer cortical surface color-coded with the curvature of its cortical foldings. Our method allows us to map back this measure and any other biomarker to the input template using the composition of the inverse deformations $\Phi^{-1}$.}
    \label{fig:application}
\end{figure}

Finally, as a by-product of CorticalFlow's deformable and diffeomorphic nature, one can seamlessly obtain a sub-voxel resolution segmentation by applying a voxelization engine. This can capture variations below the image resolution while traditional segmentation methods~\cite{roy2019quicknat} are restrained from working at the image resolution (see Figure~\ref{fig:app_highres_seg}). Additionally, an essential component of computational neuro-anatomy consists of computing local shape descriptors for different individuals and transferring them to the same reference space using conformal mappings~\cite{gu2004genus,su2015shape}. For the proposed model, one can efficiently compute the inverse transformation $\Phi^{-1}$ as shown in Figure~\ref{fig:app_curvature} for the surface curvature descriptor.

 \section{Conclusion}

This paper introduces CorticalFlow - a geometric deep learning model for efficiently reconstructing high-resolution, accurate, and regular triangular meshes from volumetric images. 
We develop a lightweight neural network to predict a dense 3D flow vector field from a volumetric image. 
Then, we describe a new Diffeomorphic Mesh Deformation (DMD) module, which is parameterized by a set of diffeomorphic mappings. This includes the derivation of numerical conditions for recasting the continuous flow ODE problem into an efficient discrete solver.
Finally, we extensively verify that the proposed model achieves state-of-the-art performance in the challenging brain cortical surface reconstruction problem.
This benchmark reveals that CorticalFlow is more accurate and, by construction, more robust to image artifacts providing anatomically plausible surfaces. 
Thanks also to its low space and time complexity, the proposed method can facilitate large-scale medical studies and support new healthcare applications.

\section{Compliance with Ethical Standards}
This research was approved by CSIRO ethics 2020 068 LR.

\section{Acknowledgements}
This work was funded in part through an Australian Department of Industry, Energy and Resources CRC-P project between CSIRO, Maxwell Plus and I-Med Radiology Network.

% we limit the space and time complexity of the resulting method, enabling its use in large-scale medical studies and new healthcare applications.

% In this paper, we introduce CorticalFlow - a geometric deep learning model for efficiently reconstructing high-resolution, accurate, and regular triangular meshes from volumetric images, supported by a theoretical framework. When applied to brain surface estimation from brain MRI, compared to state of the art existing methods, the surfaces were more accurate and presented fewer artefacts while the method had a lower memory footprint (less than 5GB for a standard MRI) and a processing time of less than 1s (much faster than the methods we tested). 

\clearpage

\vbox{%
    \hsize\textwidth
    \linewidth\hsize
    \vskip 0.1in
    \toptitlebar
    \centering
    {\LARGE\bf Supplementary Material\par}
    \bottomtitlebar}
    
\section*{Implementation details}
\hrule

\subsection*{Ablation study for Cortical Flow: number of deformation blocks}

\begin{figure}[!h]
    \centering
    \includegraphics[trim=0 0 0 50,clip,width=0.6\textwidth]{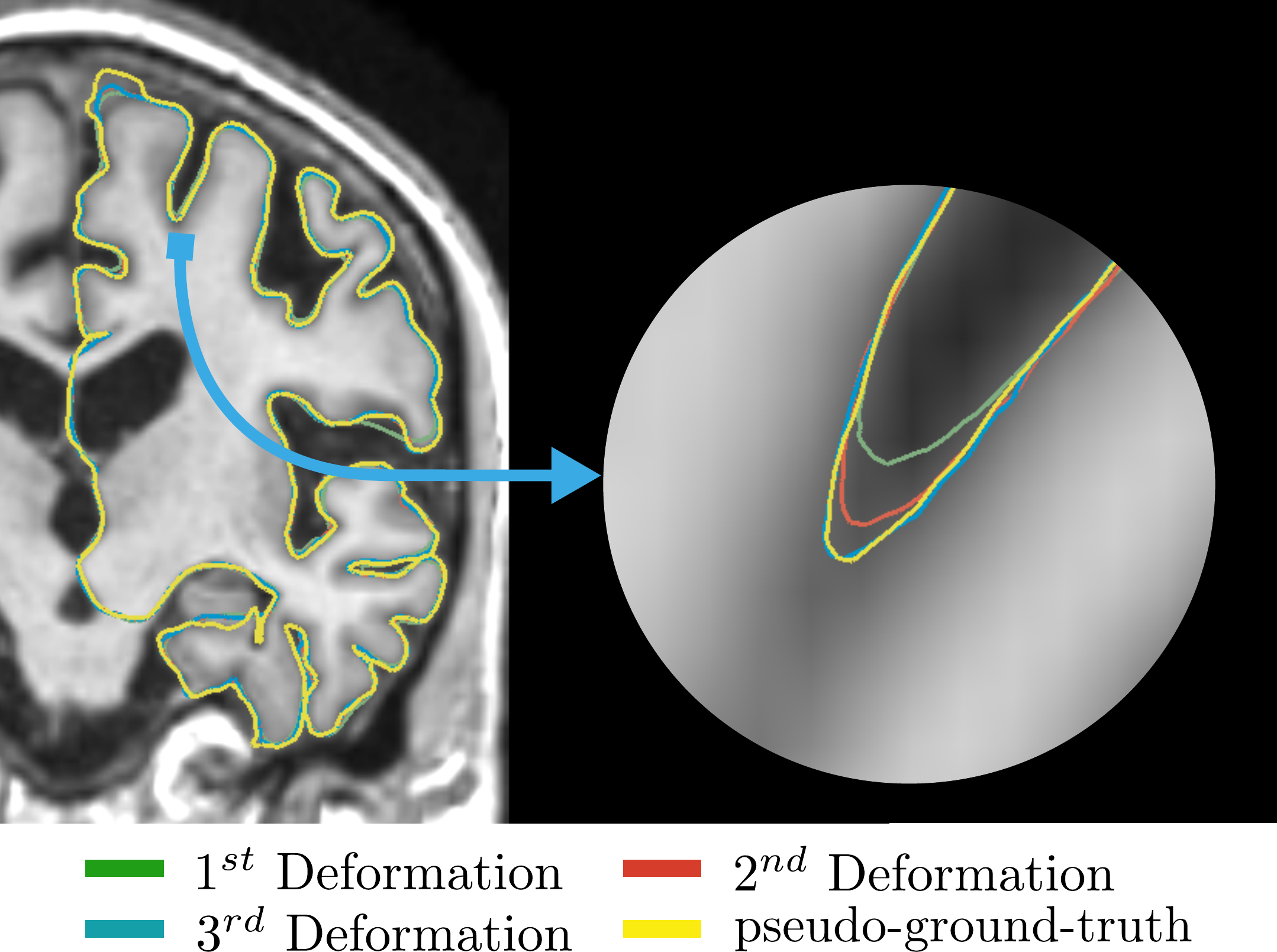}
    \caption{Example of MRI slice and outer cortical surface predicted at different deformation blocks of the proposed CorticalFlow. Zoomed in sulci region to visualize the improvements provided by the sequence of deformation predicted by CorticalFlow.}
    \label{fig:deformation_blocks}
\end{figure}

As explained in subsection~3 of our paper, CorticalFlow leverages a chain of 3 deformation blocks to provide a coarse-to-fine approximation of the targeted surface.
We evaluate CorticalFlow predictions after each deformation block in our cortical surface reconstruction benchmark to empirically validate this modeling decision.
As shown in Table~\ref{tab:number_deform}, every deformation block added allows a better approximation of the ground-truth surfaces.
More specifically, on average across all surfaces, adding a second deformation block reduces the Chamfer distance metric by 36.73\%, while adding a third deformation block reduces the same metric by a further 5.06\%. 
Importantly, this error reduction is more evident in the sulci region of the cortical surfaces, as shown by the example depicted in Figure~\ref{fig:deformation_blocks}.
\begin{table}[!htb]
\centering
\resizebox{.8\textwidth}{!}{\begin{tabular}{l|ccc|ccc|}
\hline
& \multicolumn{3}{c|}{\large{Left Outer Surface}} & \multicolumn{3}{c|}{\large{Right Outer Surface }} \\
\makecell{Number of \\ Deformation Blocks} & CH($mm$) $\downarrow$ & HD($mm$) $\downarrow$ & CHN $\uparrow$ & CH($mm$) $\downarrow$ & HD($mm$) $\downarrow$ & CHN $\uparrow$\\
\cline{1-7}
 \makecell{\large{1}} & \makecell{$1.176$ \\ $(\pm 0.143)$} & \makecell{$1.223$ \\ $(\pm 0.097)$} & \makecell{$0.903$ \\ $(\pm 0.008)$} & \makecell{$1.300$ \\ $(\pm 0.148)$} & \makecell{$1.328$ \\ $(\pm 0.103)$} & \makecell{$0.898$ \\ $(\pm 0.009)$}\\
 \makecell{\large{2}} & \makecell{$0.716$ \\ $(\pm 0.102)$} & \makecell{$0.833$ \\ $(\pm 0.056)$} & \makecell{$0.932$ \\ $(\pm 0.006)$} & \makecell{$0.741$ \\ $(\pm 0.095)$} & \makecell{$0.855$ \\ $(\pm 0.055)$} & \makecell{$0.929$ \\ $(\pm 0.006)$}\\
 \makecell{\large{3}} & \makecell{$0.681$ \\ $(\pm 0.098)$} & \makecell{$0.802$ \\ $(\pm 0.049)$} & \makecell{$0.932$ \\ $(\pm 0.006)$} & \makecell{$0.693$ \\ $(\pm 0.091)$} & \makecell{$0.815$ \\ $(\pm 0.046)$} & \makecell{$0.929$ \\ $(\pm 0.006)$}\\
\hline
& \multicolumn{3}{c|}{\large{Left Inner Surface}} & \multicolumn{3}{c|}{\large{Right Inner Surface }} \\
\makecell{Number of \\ Deformation Blocks} & CH($mm$) $\downarrow$ & HD($mm$) $\downarrow$ & CHN $\uparrow$ & CH($mm$) $\downarrow$ & HD($mm$) $\downarrow$ & CHN $\uparrow$\\
\cline{1-7}
 \makecell{\large{1}} & \makecell{$0.952$ \\ $(\pm 0.147)$} & \makecell{$1.125$ \\ $(\pm 0.133)$} & \makecell{$0.918$ \\ $(\pm 0.009)$} & \makecell{$0.917$ \\ $(\pm 0.124)$} & \makecell{$1.079$ \\ $(\pm 0.105)$} & \makecell{$0.919$ \\ $(\pm 0.009)$}\\
 \makecell{\large{2}} & \makecell{$0.636$ \\ $(\pm 0.101)$} & \makecell{$0.814$ \\ $(\pm 0.067)$} & \makecell{$0.941$ \\ $(\pm 0.007)$} & \makecell{$0.627$ \\ $(\pm 0.097)$} & \makecell{$0.811$ \\ $(\pm 0.066)$} & \makecell{$0.942$ \\ $(\pm 0.007)$}\\
 \makecell{\large{3}} & \makecell{$0.608$ \\ $(\pm 0.098)$} & \makecell{$0.785$ \\ $(\pm 0.060)$} & \makecell{$0.941$ \\ $(\pm 0.007)$} & \makecell{$0.599$ \\ $(\pm 0.093)$} & \makecell{$0.783$ \\ $(\pm 0.059)$} & \makecell{$0.942$ \\ $(\pm 0.007)$}\\
\hline
\end{tabular}}
\caption{Evaluation of the proposed CorticalFlow model with different number of deformation blocks in the cortical surface reconstruction benchmark. Consider the evaluation metrics:  chamfer distance (CH), Hausdorff distance (HD), and chamfer normals (CHN). $\downarrow$ indicates smaller metric value is better, while $\uparrow$ indicates greater metric value is better.}
\label{tab:number_deform}
\end{table}

\subsection*{Cortical Flow Architecture and Training details}
\begin{figure}[!h]
    \centering
    \includegraphics[width=\textwidth]{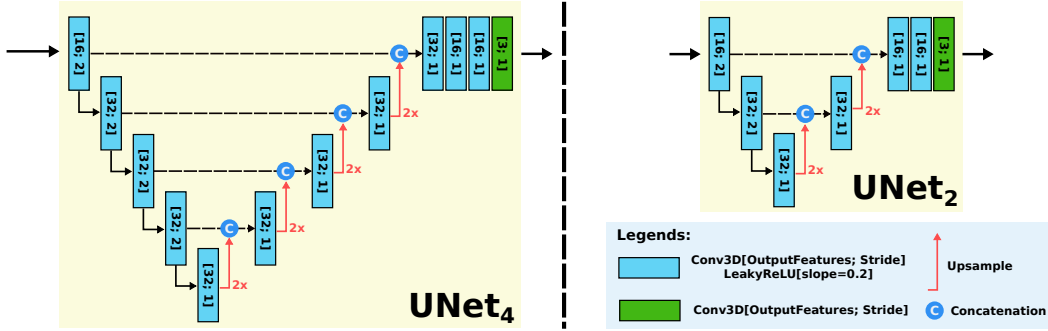}
    \caption{UNet architectures used in CorticalFlow. See Figure~2 of our paper for a global overview of the model.}
    \label{fig:unet_archs}
\end{figure}

As shown in Figure~2 of our paper, CorticalFlow consists of a chain of three deformation blocks. 
Each of these deformations is implemented by some UNet flow vector field predictor and the Diffeomorphic Mesh Deformation (DMD) module described in subsection~3.1 of our paper.
More specifically, the first deformation block uses our $\textbf{UNet}_{\textbf{4}}$ architecture while the remaining ones use our $\textbf{UNet}_{\textbf{2}}$ architecture. 
Both architectures are described in details in Figure~\ref{fig:unet_archs}. As explained in subsection~3.2, the reason for this architectural change is to promote the learning of a coarse-to-fine sequence of deformation blocks. 

For training CorticalFlow, we adopted a sequential approach where we train one deformation block at a time while freezing the weights of the previous UNet(s). 
All deformation blocks are trained according to equation~2 for 70k iterations with a batch size of three image-surface pairs. 
As training loss $\mathcal{L}(\cdot, \cdot)$, we minimize the mesh edge loss and Chamfer distance computed on point clouds of 150k points sampled from the predicted and ground-truth surfaces using random uniform sampling. The implementation of these losses and sampling algorithm are provided in the \texttt{PyTorch3D} library~\citep{ravi2020pytorch3d}. 
As an optimizer for each deformation, we use Adam~\citep{kingma2014adam} with an initial learning rate of $10^{-4}$. 
Both predicted and ground-truth surfaces are shrunk to lie in the unit ball to normalize the learning loss. 

\subsection*{Backbone selection for Neural Mesh Flow (NMF)}
\begin{figure}[!ht]
    \centering
    \includegraphics[width=\textwidth,trim={0 0 70cm 0},clip]{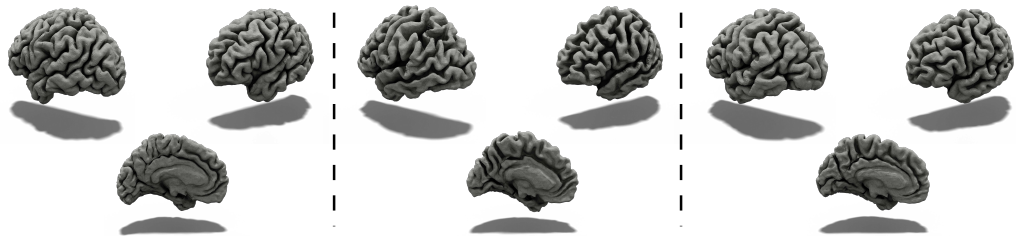}
    \caption{Outer cortical surface reconstruction for four subjects (columns) using the NMF framework with different backbones. The Hypercolumns backbone is significantly better than the others, and thus it is the NMF's backbone presented in our benchmark.}
    \label{fig:nmf_backbones}
\end{figure}

Neural Mesh Flow (NMF) is a deformation-based geometric deep learning model for retrieving regular surfaces for objects depicted in a single 2D image.
This model has two main blocks: an image-level feature encoder and a mesh transformer.
The former is composed of a ResNet~\citep{he2016deep} point-cloud predictor and a PointNet~\citep{qi2017pointnet} network providing an image-level feature vector representation for an input 2D image. 
The latter receives this image-level feature vector as input and deforms a prescribed template towards the ground-truth surface using Neural Ordinary Differential Equation (NODE) blocks.
To adapt this model to cortical surface reconstruction performing only minimal changes, we swap the architecture of the point-cloud predictor from a 2D ResNet to a 3D ResNet. 
However, as shown in the first row of Figure~\ref{fig:nmf_backbones}, the resulting model performs very poorly in the cortical surface reconstruction task. 
More specifically, We found it very hard to predict point clouds to cortical surfaces since these surfaces present many local dissimilarities (e.g., cortical folding patterns) that are hard to capture by funnel-like architectures like ResNet. 
Trying to overcome this problem, we replaced the ResNet with a 3D UNet~\citep{ronneberger2015u} with shortcut connections (i.e., our Unet$_4$ architecture) to exploit high-resolution feature maps within the computation of the image-level feature vector representation. 
As shown in the second row of Figure~\ref{fig:nmf_backbones}, the results were still far from satisfactory. 
As discussed in subsection 2.2 and also observed by \citet{cruz2021deepcsr}, an image-level feature vector does not hold fine-grained information enough to reconstruct cortical surfaces accurately.
Therefore, we tried their proposed Hypercolumns architecture which equips each template vertex with a local feature descriptor of the input image resulting in more accurate cortical surface reconstructions as shown in the last row of Figure~\ref{fig:nmf_backbones}. 
Therefore, the NMF with Hypercolumns backbone is the NMF model used in our benchmark, whose results are summarized in Table~1 of our paper.

\subsection*{QuickNAT Baseline}
\begin{figure}[!h]
    \centering
    \includegraphics[width=\textwidth]{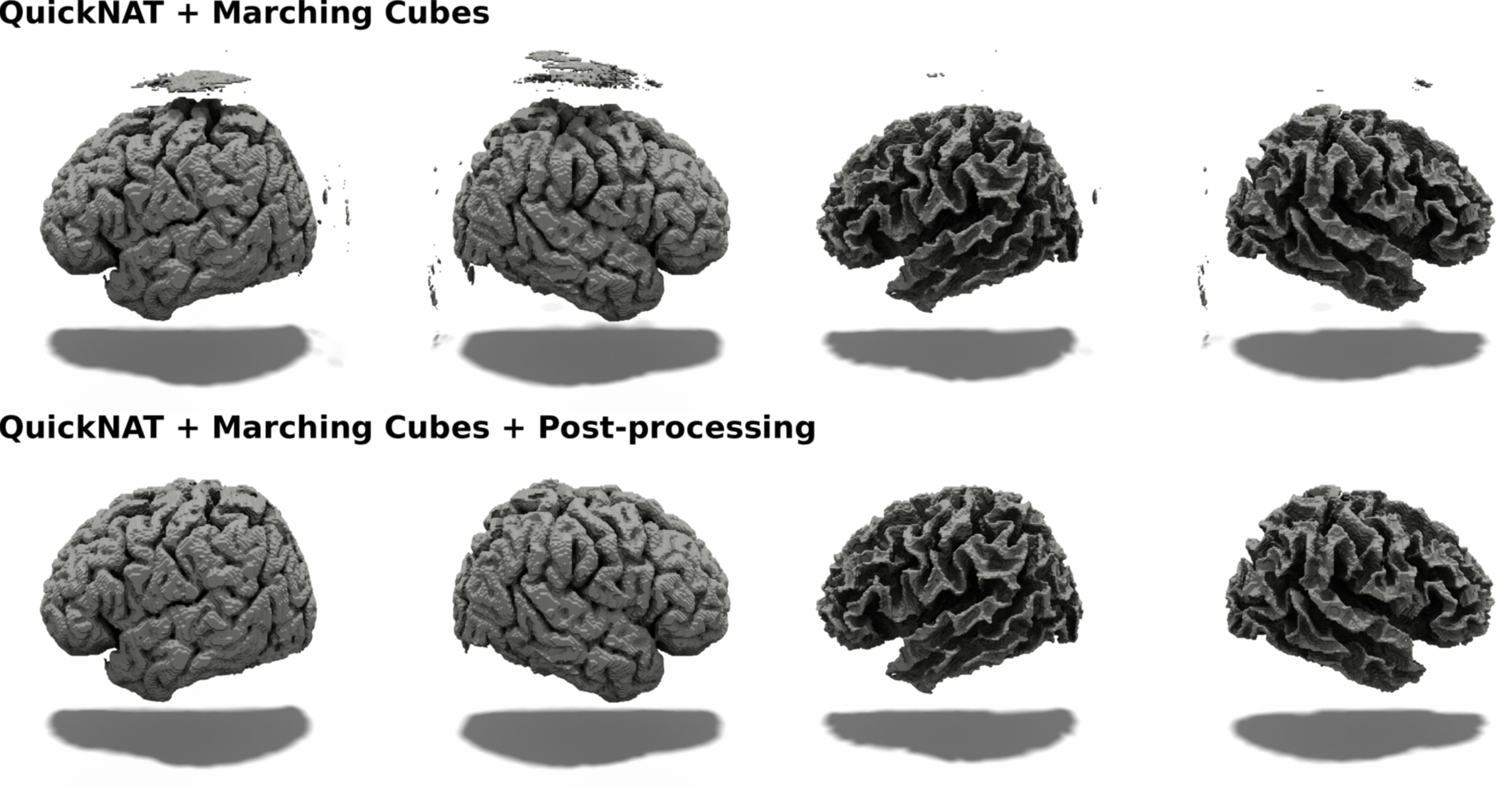}
    \caption{Cortical Surfaces generated with the QuickNAT baseline before and after the post-processing routine.}
    \label{fig:quicknat_meshes}
\end{figure}

The QuickNAT baseline consists of a voxel-wise segmentation model, iso-surface extraction method, and a mesh post-processing routine. 
More specifically, we first predict a segmentation of the input MRI into 28 anatomical regions using the QuickNAT~\citep{roy2019quicknat} state-of-the-art segmentation model for brain segmentation.
Second, we build the four cortical volumes by assembling anatomical structures contained in the four surfaces. 
Third, we run a marching cubes~\citep{lewiner2003efficient} algorithm to retrieve triangle meshes from the obtained binary segmentations. 
Since the resulting meshes present a lot of unwanted connected components (due to spurious mistakes in the segmentation), we only isolate the largest connected component using the \texttt{trimesh.graph.connected\_component\_labels} function.
Figure~\ref{fig:quicknat_meshes} presents some meshes generated with this QuickNAT baseline.
Our methodology was to get the best geometrical measure stemming from a segmentation-based approach. We verify that the suppression of small connected components improves all the metrics presented in our paper.
Note, however, that those meshes cannot be used for the cortical surface reconstruction problem since they comprise hole and handle (not 0-genus). It is also important to notice that numerous errors are imputable to the limited resolution of this approach.

\subsection*{Dataset Information and Preprocessing}
The dataset used in the experiments described in subsection~4 of our paper was introduced in \citep{cruz2021deepcsr}.
It consists of MR images extracted from the Alzheimer’s Disease Neuroimaging Initiative\footnote{Data used in preparation of this article were obtained from the Alzheimer’s Disease Neuroimaging Initiative
(ADNI) database (adni.loni.usc.edu). As such, the investigators within the ADNI contributed to the design
and implementation of ADNI and/or provided data but did not participate in analysis or writing of this report.
A complete listing of ADNI investigators can be found at:
\url{http://adni.loni.usc.edu/wp-content/uploads/how_to_apply/ADNI_Acknowledgement_List.pdf}} (ADNI) \citep{Jack2008:ADNI} and their respective pseudo-ground-truth surfaces generated with the FreeSurfer V6.0 cross-subsectional pipeline~\citep{fischl2002whole,fischl1999cortical}.
It comprises 3876 MR images from 820 different subjects collected at different time points and their respective cortical surfaces split by brain hemisphere (i.e., left outer surface, left inner surface, right outer surface, and right inner surface).
This dataset is split by subjects resulting in 2353 MRI scans from 492 subjects for training ($\approx 60\%$), 375 MRI scans from 82 subjects for validation ($\approx 10\%$), and 1148 MRI scans from 246 subjects for testing ($\approx 30\%$).  
It is also important to emphasize that these splits do not have MRIs or subjects in common for an unbiased evaluation.

As preprocessing, the original ADNI images are conformed and normalized according to the first steps in the FreeSurfer V6 pipeline. 
These images are saved at \path{<subject_id>/mri/orig.mgz} on the FreeSurfer output directory. 
Then, they are affine registered to the MNI105 brain template \citep{mazziotta1995probabilistic} using the \texttt{NiftyReg} toolbox \citep{modat2014global}. 
Their respective pseudo-ground-truth surfaces are also transformed using the computed transformation. 
Finally, for memory efficiency, these images are split by hemisphere since we learn a model for each surface resulting in 1$mm^2$ isotropic T1-weighted images with $96\times192\times160$ dimensions. 
Detailed instructions for downloading and preprocessing this data will be provided along with our source code.

\newpage
\section*{Proof and discussion}
\hrule

\subsection*{Proof of Theorem~3.2.}

\begin{customthm}{3.2}
\label{EDO}
Existence and uniqueness of the solution. Define $\Phi$ through the autonomous ODE,
\begin{equation}\label{eqn:autoODE}
\frac{\text{d}\Phi(s;\mathbf{x})}{\text{d}s} = v\left(\Phi(s;\mathbf{x})\right), \text{ with } \Phi(0;\mathbf{x}) = \mathbf{x}.
\end{equation}
Then $\Phi$ is uniquely defined on $\mathbb{R}\times \Omega$, is Lipschitz and for each $t$, the mapping $x\mapsto \Phi(t,x)$ is bijective with Lipschitz inverse.
\end{customthm}

\begin{proof}
This is a standard result in flow theory; see Theorem 1.2.6~\citet{berger2012differential}.

\begin{enumerate}
\item First, notice that the vector field $v$ is L-Lipschitz with $L > 0$ defined as
\[
L = \text{Lip}(v)  = \max_{i \in \{1,2,3\}} \frac{\| \nabla_{i} \mathbf{U} \|_{2,\infty}}{d_i}
\]
with $\nabla_{d} \mathbf{U}$ the forward first order finite difference operator in the $d$-th direction with zero padding
\[
\nabla_1 U_{i,j,k} = \begin{cases} \text{if  } i \in \llbracket 1,n-1 \rrbracket, \quad  U_{i+1,j,k} - U_{i,j,k} \\
\text{if  } i \in \{0,n\}, \quad U_{i,j,k}.
\end{cases}
\]

\item Then verify that vector field $v$ is bounded by $M$ with $ M = \| v \|_{2,\infty} = \max_{i,j,k} \| \mathbf{U}_{ijk} \|_2$

\end{enumerate}

With these two constants one can use the result of~\citet{berger2012differential} (Theorem 1.2.6) to define a unique local solution for $t \in ]-b,b[$ with $b < \inf(\frac{\text{diam}(\Omega)}{M},\frac{1}{L})$ where $\text{diam}(\Omega)$ is the diameter of the set $\Omega$. 

To extend this result for all $t \in \R$, one has to notice that the solution is defined for every $t$ since the integral solutions are contained in $\Omega$ since $v=0$ on $\partial \Omega$.
To construct the inverse, one uses the same proof method but integrates $v$ from $t$ to zero.
\end{proof}

\subsection*{Caveat: Discrete approximation of continuous surfaces}

\begin{figure}[!h]
\centering
\includegraphics[height=2cm]{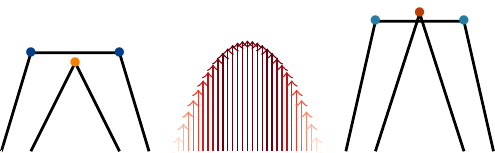}\hspace*{.3cm}
\includegraphics[trim=10 10 10 5,clip,height=2cm]{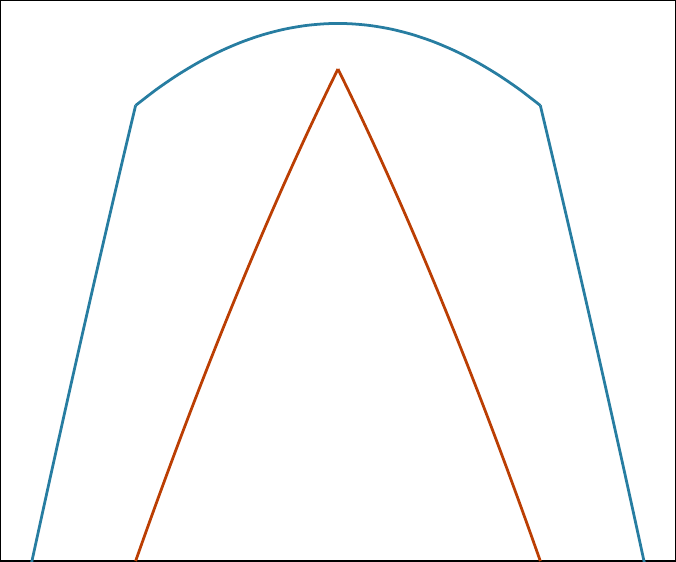}
\caption{Deformation of a template mesh. From left to right: The original triangle mesh, a prescribed vector-field $v$, deformed triangulation where the map $\Psi$ applied on the mesh's vertices creating a self-intersection, and continuous deformation of triangle mesh' surface by $\Psi$.}\label{fig:caveat}
\end{figure}

As explained at the end of Section~3.1, $\Psi$ being a homeomorphism does not guarantee that the discrete problem we are solving is immune to self-intersection, and such a pathological case is described in Figure~\ref{fig:caveat}. Note, however, that one could get rid of the self-intersection up to sufficient remeshing of the self-intersecting faces.
This discretization issue has to be kept in mind when using a deformable model that acts on the vertices of a triangle mesh.

\clearpage
\section*{Further comparison with pre-existing methods}
\hrule
\subsection*{Comparison with DeepCSR and NMF}
\begin{figure}[!ht]
    \centering
    \includegraphics[width=\textwidth]{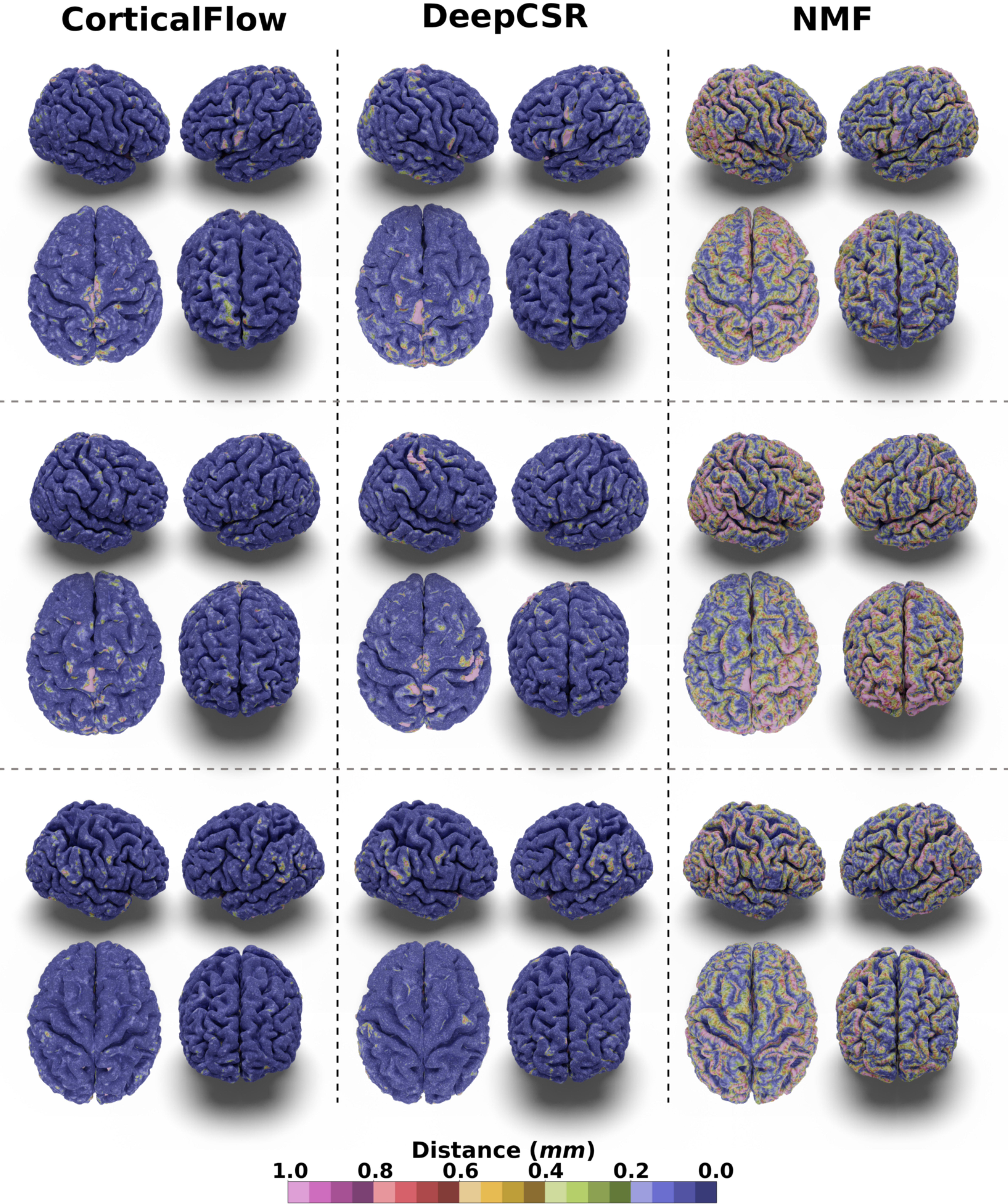}
    \caption{More examples of predicted \textbf{outer cortical surfaces} color-coded with the distance to the pseudo-ground-truth surfaces as shown in Figure~3 of our paper. Here, each row presents the results for a different input MRI. All the presented anatomies are included between the $40$th and $60$th percentile for the Chamfer distance.}
    \label{fig:ch_colorcoded_pial}
\end{figure}

\begin{figure}[!htb]
    \centering
    \includegraphics[width=\textwidth]{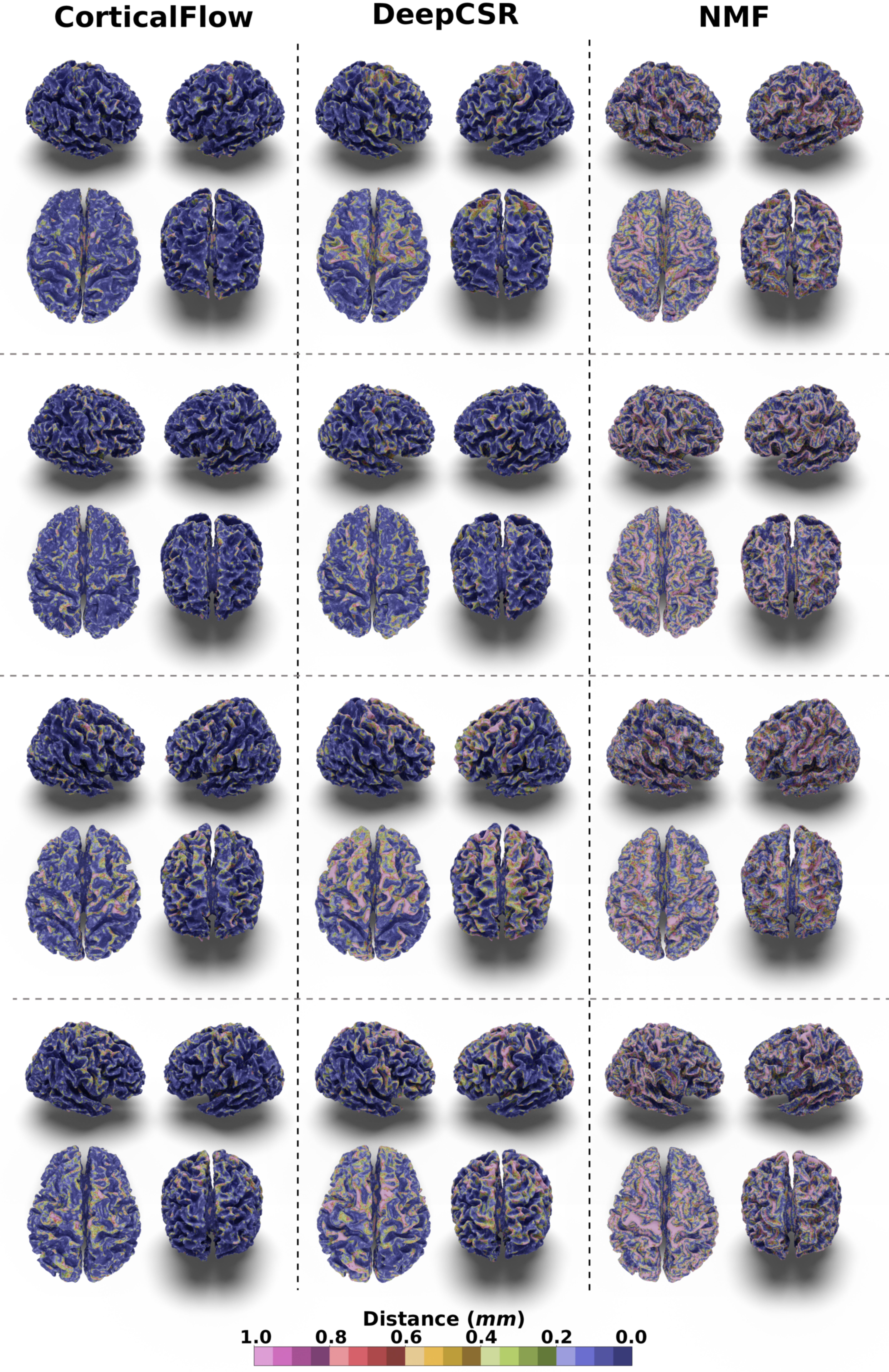}
    \caption{More examples of predicted \textbf{inner cortical surfaces} color-coded with the distance to the pseudo-ground-truth surfaces as shown in Figure~3 of our paper. Here, each row presents the results for a different input MRI. All the presented anatomies are included between the $40$th and $60$th percentile for the Chamfer distance.}
    \label{fig:ch_colorcoded_white}
\end{figure}

\clearpage

\bibliographystyle{plainnat}
\bibliography{nips_biblio}

\begin{thebibliography}{81}
\providecommand{\natexlab}[1]{#1}
\providecommand{\url}[1]{\texttt{#1}}
\expandafter\ifx\csname urlstyle\endcsname\relax
  \providecommand{\doi}[1]{doi: #1}\else
  \providecommand{\doi}{doi: \begingroup \urlstyle{rm}\Url}\fi

\bibitem[Arsigny(2004)]{arsigny2004processing}
Vincent Arsigny.
\newblock \emph{Processing data in lie groups: an algebraic approach.
  application to non-linear registration and diffusion tensor mri}.
\newblock PhD thesis, Citeseer, 2004.

\bibitem[Arsigny et~al.(2005)Arsigny, Pennec, and Ayache]{arsigny2005polyrigid}
Vincent Arsigny, Xavier Pennec, and Nicholas Ayache.
\newblock Polyrigid and polyaffine transformations: a novel geometrical tool to
  deal with non-rigid deformations--application to the registration of
  histological slices.
\newblock \emph{Medical image analysis}, 9\penalty0 (6):\penalty0 507--523,
  2005.

\bibitem[Ashburner(2007)]{ashburner2007fast}
John Ashburner.
\newblock A fast diffeomorphic image registration algorithm.
\newblock \emph{Neuroimage}, 38\penalty0 (1):\penalty0 95--113, 2007.

\bibitem[Bazin and Pham(2007)]{Bazin:2007}
Pierre-Louis Bazin and Dzung~L Pham.
\newblock Topology correction of segmented medical images using a fast marching
  algorithm.
\newblock \emph{Computer methods and programs in biomedicine}, 88\penalty0
  (2):\penalty0 182--190, 2007.

\bibitem[Beg et~al.(2005)Beg, Miller, Trouv{\'e}, and Younes]{beg2005computing}
M~Faisal Beg, Michael~I Miller, Alain Trouv{\'e}, and Laurent Younes.
\newblock Computing large deformation metric mappings via geodesic flows of
  diffeomorphisms.
\newblock \emph{International journal of computer vision}, 61\penalty0
  (2):\penalty0 139--157, 2005.

\bibitem[Berger and Gostiaux(2012)]{berger2012differential}
Marcel Berger and Bernard Gostiaux.
\newblock \emph{Differential Geometry: Manifolds, Curves, and Surfaces:
  Manifolds, Curves, and Surfaces}, volume 115.
\newblock Springer Science \& Business Media, 2012.

\bibitem[Camion and Younes(2001)]{camion2001geodesic}
Vincent Camion and Laurent Younes.
\newblock Geodesic interpolating splines.
\newblock In \emph{International workshop on energy minimization methods in
  computer vision and pattern recognition}, pages 513--527. Springer, 2001.

\bibitem[Charon(2013)]{charon2013analysis}
Nicolas Charon.
\newblock \emph{Analysis of geometric and functional shapes with extensions of
  currents: applications to registration and atlas estimation}.
\newblock PhD thesis, {\'E}cole normale sup{\'e}rieure de Cachan-ENS Cachan,
  2013.

\bibitem[Chen et~al.(2018)Chen, Rubanova, Bettencourt, and
  Duvenaud]{chen2018neuralode}
Ricky T.~Q. Chen, Yulia Rubanova, Jesse Bettencourt, and David Duvenaud.
\newblock Neural ordinary differential equations.
\newblock \emph{Advances in Neural Information Processing Systems}, 2018.

\bibitem[Chen et~al.(2020)Chen, Tagliasacchi, and Zhang]{chen2020bsp}
Zhiqin Chen, Andrea Tagliasacchi, and Hao Zhang.
\newblock Bsp-net: Generating compact meshes via binary space partitioning.
\newblock In \emph{Proceedings of the IEEE/CVF Conference on Computer Vision
  and Pattern Recognition}, pages 45--54, 2020.

\bibitem[Choy et~al.(2016)Choy, Xu, Gwak, Chen, and Savarese]{choy20163d}
Christopher~B Choy, Danfei Xu, JunYoung Gwak, Kevin Chen, and Silvio Savarese.
\newblock 3d-r2n2: A unified approach for single and multi-view 3d object
  reconstruction.
\newblock In \emph{European conference on computer vision}, pages 628--644.
  Springer, 2016.

\bibitem[Cignoni et~al.(2008)Cignoni, Callieri, Corsini, Dellepiane, Ganovelli,
  and Ranzuglia]{LocalChapterEvents:ItalChap:ItalianChapConf2008:129-136}
Paolo Cignoni, Marco Callieri, Massimiliano Corsini, Matteo Dellepiane, Fabio
  Ganovelli, and Guido Ranzuglia.
\newblock {MeshLab: an Open-Source Mesh Processing Tool}.
\newblock In Vittorio Scarano, Rosario~De Chiara, and Ugo Erra, editors,
  \emph{Eurographics Italian Chapter Conference}. The Eurographics Association,
  2008.
\newblock ISBN 978-3-905673-68-5.
\newblock
  \doi{10.2312/LocalChapterEvents/ItalChap/ItalianChapConf2008/129-136}.

\bibitem[Cooley et~al.(2020)Cooley, McDaniel, Stockmann, Srinivas, Cauley,
  Sliwiak, Sappo, Vaughn, Guerin, Rosen, et~al.]{cooley2020portable}
Clarissa~Z Cooley, Patrick~C McDaniel, Jason~P Stockmann, Sai~Abitha Srinivas,
  Stephen Cauley, Monika Sliwiak, Charlotte~R Sappo, Christopher~F Vaughn,
  Bastien Guerin, Matthew~S Rosen, et~al.
\newblock A portable brain mri scanner for underserved settings and
  point-of-care imaging.
\newblock \emph{arXiv preprint arXiv:2004.13183}, 2020.

\bibitem[Dahnke et~al.(2013)Dahnke, Yotter, and Gaser]{Dahnke:NI2013:CAT}
Robert Dahnke, Rachel~Aine Yotter, and Christian Gaser.
\newblock Cortical thickness and central surface estimation.
\newblock \emph{NeuroImage}, 65:\penalty0 336--348, 2013.

\bibitem[Dalca et~al.(2019)Dalca, Balakrishnan, Guttag, and
  Sabuncu]{dalca2019unsupervised}
Adrian~V Dalca, Guha Balakrishnan, John Guttag, and Mert~R Sabuncu.
\newblock Unsupervised learning of probabilistic diffeomorphic registration for
  images and surfaces.
\newblock \emph{Medical image analysis}, 57:\penalty0 226--236, 2019.

\bibitem[Dale et~al.(1999)Dale, Fischl, and Sereno]{Dale:NI1999}
Anders~M Dale, Bruce Fischl, and Martin~I Sereno.
\newblock Cortical surface-based analysis: I. segmentation and surface
  reconstruction.
\newblock \emph{NeuroImage}, 9\penalty0 (2):\penalty0 179--194, 1999.

\bibitem[Deng et~al.(2020)Deng, Genova, Yazdani, Bouaziz, Hinton, and
  Tagliasacchi]{deng2020cvxnet}
Boyang Deng, Kyle Genova, Soroosh Yazdani, Sofien Bouaziz, Geoffrey Hinton, and
  Andrea Tagliasacchi.
\newblock Cvxnet: Learnable convex decomposition.
\newblock In \emph{Proceedings of the IEEE/CVF Conference on Computer Vision
  and Pattern Recognition}, pages 31--44, 2020.

\bibitem[Du et~al.(2007)Du, Schuff, Kramer, Rosen, Gorno-Tempini, Rankin,
  Miller, and Weiner]{du2007different}
An-Tao Du, Norbert Schuff, Joel~H Kramer, Howard~J Rosen, Maria~Luisa
  Gorno-Tempini, Katherine Rankin, Bruce~L Miller, and Michael~W Weiner.
\newblock Different regional patterns of cortical thinning in alzheimer's
  disease and frontotemporal dementia.
\newblock \emph{Brain}, 130\penalty0 (4):\penalty0 1159--1166, 2007.

\bibitem[Dupuis et~al.(1998)Dupuis, Grenander, and
  Miller]{dupuis1998variational}
Paul Dupuis, Ulf Grenander, and Michael~I Miller.
\newblock Variational problems on flows of diffeomorphisms for image matching.
\newblock \emph{Quarterly of applied mathematics}, pages 587--600, 1998.

\bibitem[Ebin and Marsden(1970)]{ebin1970groups}
David~G Ebin and Jerrold Marsden.
\newblock Groups of diffeomorphisms and the motion of an incompressible fluid.
\newblock \emph{Annals of Mathematics}, pages 102--163, 1970.

\bibitem[Engwirda and Ivers(2016)]{engwirda2016off}
Darren Engwirda and David Ivers.
\newblock Off-centre steiner points for delaunay-refinement on curved surfaces.
\newblock \emph{Computer-Aided Design}, 72:\penalty0 157--171, 2016.

\bibitem[Fahim et~al.(2021)Fahim, Amin, and Zarif]{fahim2021single}
George Fahim, Khalid Amin, and Sameh Zarif.
\newblock Single-view 3d reconstruction: A survey of deep learning methods.
\newblock \emph{Computers \& Graphics}, 94:\penalty0 164--190, 2021.

\bibitem[Feydy(2020)]{feydy2020analyse}
Jean Feydy.
\newblock \emph{Analyse de donn{\'e}es g{\'e}om{\'e}triques, au del{\`a} des
  convolutions}.
\newblock PhD thesis, Universit{\'e} Paris-Saclay, 2020.

\bibitem[Fischl et~al.(1999)Fischl, Sereno, and Dale]{fischl1999cortical}
Bruce Fischl, Martin~I Sereno, and Anders~M Dale.
\newblock Cortical surface-based analysis: Ii: inflation, flattening, and a
  surface-based coordinate system.
\newblock \emph{Neuroimage}, 9\penalty0 (2):\penalty0 195--207, 1999.

\bibitem[Fischl et~al.(2002)Fischl, Salat, Busa, Albert, Dieterich, Haselgrove,
  Van Der~Kouwe, Killiany, Kennedy, Klaveness, et~al.]{fischl2002whole}
Bruce Fischl, David~H Salat, Evelina Busa, Marilyn Albert, Megan Dieterich,
  Christian Haselgrove, Andre Van Der~Kouwe, Ron Killiany, David Kennedy, Shuna
  Klaveness, et~al.
\newblock Whole brain segmentation: automated labeling of neuroanatomical
  structures in the human brain.
\newblock \emph{Neuron}, 33\penalty0 (3):\penalty0 341--355, 2002.

\bibitem[Genova et~al.(2019)Genova, Cole, Vlasic, Sarna, Freeman, and
  Funkhouser]{genova2019learning}
Kyle Genova, Forrester Cole, Daniel Vlasic, Aaron Sarna, William~T Freeman, and
  Thomas Funkhouser.
\newblock Learning shape templates with structured implicit functions.
\newblock In \emph{Proceedings of the IEEE/CVF International Conference on
  Computer Vision}, pages 7154--7164, 2019.

\bibitem[Girdhar et~al.(2016)Girdhar, Fouhey, Rodriguez, and
  Gupta]{girdhar2016learning}
Rohit Girdhar, David~F Fouhey, Mikel Rodriguez, and Abhinav Gupta.
\newblock Learning a predictable and generative vector representation for
  objects.
\newblock In \emph{European Conference on Computer Vision}, pages 484--499.
  Springer, 2016.

\bibitem[Groueix et~al.(2018)Groueix, Fisher, Kim, Russell, and
  Aubry]{groueix2018papier}
Thibault Groueix, Matthew Fisher, Vladimir~G Kim, Bryan~C Russell, and Mathieu
  Aubry.
\newblock A papier-m{\^a}ch{\'e} approach to learning 3d surface generation.
\newblock In \emph{Proceedings of the IEEE conference on computer vision and
  pattern recognition}, pages 216--224, 2018.

\bibitem[Gu et~al.(2004)Gu, Wang, Chan, Thompson, and Yau]{gu2004genus}
Xianfeng Gu, Yalin Wang, Tony~F Chan, Paul~M Thompson, and Shing-Tung Yau.
\newblock Genus zero surface conformal mapping and its application to brain
  surface mapping.
\newblock \emph{IEEE transactions on medical imaging}, 23\penalty0
  (8):\penalty0 949--958, 2004.

\bibitem[Gupta and Chandraker(2020)]{gupta_neurips20_nmf}
Kunal Gupta and Manmohan Chandraker.
\newblock Neural mesh flow: 3d manifold mesh generation via diffeomorphic
  flows.
\newblock In \emph{Advances in Neural Information Processing Systems},
  volume~33, pages 1747--1758, 2020.

\bibitem[Han et~al.(2004)Han, Pham, Tosun, Rettmann, Xu, and
  Prince]{Han2004:NI2004:CRUISE}
Xiao Han, Dzung~L Pham, Duygu Tosun, Maryam~E Rettmann, Chenyang Xu, and
  Jerry~L Prince.
\newblock Cruise: cortical reconstruction using implicit surface evolution.
\newblock \emph{NeuroImage}, 23\penalty0 (3):\penalty0 997--1012, 2004.

\bibitem[Hanocka et~al.(2020)Hanocka, Metzer, Giryes, and
  Cohen-Or]{Hanocka2020p2m}
Rana Hanocka, Gal Metzer, Raja Giryes, and Daniel Cohen-Or.
\newblock Point2mesh: A self-prior for deformable meshes.
\newblock \emph{ACM Trans. Graph.}, 39\penalty0 (4), 2020.

\bibitem[He et~al.(2016)He, Zhang, Ren, and Sun]{he2016deep}
Kaiming He, Xiangyu Zhang, Shaoqing Ren, and Jian Sun.
\newblock Deep residual learning for image recognition.
\newblock In \emph{Proceedings of the IEEE conference on computer vision and
  pattern recognition}, pages 770--778, 2016.

\bibitem[Henschel et~al.(2020)Henschel, Conjeti, Estrada, Diers, Fischl, and
  Reuter]{Henschel:NI2020}
Leonie Henschel, Sailesh Conjeti, Santiago Estrada, Kersten Diers, Bruce
  Fischl, and Martin Reuter.
\newblock Fastsurfer-a fast and accurate deep learning based neuroimaging
  pipeline.
\newblock \emph{NeuroImage}, page 117012, 2020.

\bibitem[Häne et~al.(2017)Häne, Tulsiani, and Malik]{Hane2017}
Christian Häne, Shubham Tulsiani, and Jitendra Malik.
\newblock Hierarchical surface prediction for 3d object reconstruction.
\newblock In \emph{2017 International Conference on 3D Vision (3DV)}, pages
  412--420, 2017.
\newblock \doi{10.1109/3DV.2017.00054}.

\bibitem[Jack~Jr et~al.(2008)Jack~Jr, Bernstein, Fox, Thompson, Alexander,
  Harvey, Borowski, Britson, L.~Whitwell, Ward, et~al.]{Jack2008:ADNI}
Clifford~R Jack~Jr, Matt~A Bernstein, Nick~C Fox, Paul Thompson, Gene
  Alexander, Danielle Harvey, Bret Borowski, Paula~J Britson, Jennifer
  L.~Whitwell, Chadwick Ward, et~al.
\newblock The alzheimer's disease neuroimaging initiative (adni): Mri methods.
\newblock \emph{Journal of Magnetic Resonance Imaging}, 27\penalty0
  (4):\penalty0 685--691, 2008.

\bibitem[Jiang et~al.(2020)Jiang, Huang, Tagliasacchi, Guibas,
  et~al.]{jiang2020shapeflow}
Chiyu Jiang, Jingwei Huang, Andrea Tagliasacchi, Leonidas Guibas, et~al.
\newblock Shapeflow: Learnable deformations among 3d shapes.
\newblock \emph{arXiv preprint arXiv:2006.07982}, 2020.

\bibitem[Kim et~al.(2005)Kim, Singh, Lee, Lerch, Ad-Dab'bagh, MacDonald, Lee,
  Kim, and Evans]{Kim:NI2005:CLASP}
June~Sic Kim, Vivek Singh, Jun~Ki Lee, Jason Lerch, Yasser Ad-Dab'bagh, David
  MacDonald, Jong~Min Lee, Sun~I Kim, and Alan~C Evans.
\newblock Automated 3-d extraction and evaluation of the inner and outer
  cortical surfaces using a laplacian map and partial volume effect
  classification.
\newblock \emph{NeuroImage}, 27\penalty0 (1):\penalty0 210--221, 2005.

\bibitem[Kingma and Ba(2014)]{kingma2014adam}
Diederik~P Kingma and Jimmy Ba.
\newblock Adam: A method for stochastic optimization.
\newblock \emph{arXiv preprint arXiv:1412.6980}, 2014.

\bibitem[Koumoutsakos et~al.(2008)Koumoutsakos, Cottet, and
  Rossinelli]{koumoutsakos2008flow}
Petros Koumoutsakos, Georges-Henri Cottet, and Diego Rossinelli.
\newblock Flow simulations using particles-bridging computer graphics and cfd.
\newblock In \emph{SIGGRAPH 2008-35th International Conference on Computer
  Graphics and Interactive Techniques}, pages 1--73. ACM, 2008.

\bibitem[Krebs et~al.(2019)Krebs, Delingette, Mailh{\'e}, Ayache, and
  Mansi]{krebs2019learning}
Julian Krebs, Herv{\'e} Delingette, Boris Mailh{\'e}, Nicholas Ayache, and
  Tommaso Mansi.
\newblock Learning a probabilistic model for diffeomorphic registration.
\newblock \emph{IEEE transactions on medical imaging}, 38\penalty0
  (9):\penalty0 2165--2176, 2019.

\bibitem[Lewiner et~al.(2003)Lewiner, Lopes, Vieira, and
  Tavares]{lewiner2003efficient}
Thomas Lewiner, H{\'e}lio Lopes, Ant{\^o}nio~Wilson Vieira, and Geovan Tavares.
\newblock Efficient implementation of marching cubes' cases with topological
  guarantees.
\newblock \emph{Journal of graphics tools}, 8\penalty0 (2):\penalty0 1--15,
  2003.

\bibitem[Liu et~al.(2020)Liu, Zhang, and Su]{liu2020meshing}
Minghua Liu, Xiaoshuai Zhang, and Hao Su.
\newblock Meshing point clouds with predicted intrinsic-extrinsic ratio
  guidance.
\newblock In \emph{European Conference on Computer Vision}, pages 68--84.
  Springer, 2020.

\bibitem[Lorenzi and Pennec(2013)]{lorenzi2013geodesics}
Marco Lorenzi and Xavier Pennec.
\newblock Geodesics, parallel transport \& one-parameter subgroups for
  diffeomorphic image registration.
\newblock \emph{International journal of computer vision}, 105\penalty0
  (2):\penalty0 111--127, 2013.

\bibitem[Mangin et~al.(2004)Mangin, Riviere, Cachia, Duchesnay, Cointepas,
  Papadopoulos-Orfanos, Scifo, Ochiai, Brunelle, and
  R{\'e}gis]{mangin2004framework}
J-F Mangin, Denis Riviere, Arnaud Cachia, Edouard Duchesnay, Yves Cointepas,
  Dimitri Papadopoulos-Orfanos, Paola Scifo, T~Ochiai, Francis Brunelle, and
  Jean R{\'e}gis.
\newblock A framework to study the cortical folding patterns.
\newblock \emph{Neuroimage}, 23:\penalty0 S129--S138, 2004.

\bibitem[Mazziotta et~al.(1995)Mazziotta, Toga, Evans, Fox, Lancaster,
  et~al.]{mazziotta1995probabilistic}
John~C Mazziotta, Arthur~W Toga, Alan Evans, Peter Fox, Jack Lancaster, et~al.
\newblock A probabilistic atlas of the human brain: theory and rationale for
  its development.
\newblock \emph{Neuroimage}, 2\penalty0 (2):\penalty0 89--101, 1995.

\bibitem[Mescheder et~al.(2019)Mescheder, Oechsle, Niemeyer, Nowozin, and
  Geiger]{Mescheder_2019_CVPR}
Lars Mescheder, Michael Oechsle, Michael Niemeyer, Sebastian Nowozin, and
  Andreas Geiger.
\newblock Occupancy networks: Learning 3d reconstruction in function space.
\newblock In \emph{Proceedings of the IEEE/CVF Conference on Computer Vision
  and Pattern Recognition (CVPR)}, June 2019.

\bibitem[Michalkiewicz et~al.(2019)Michalkiewicz, Pontes, Jack, Baktashmotlagh,
  and Eriksson]{michalkiewicz2019implicit}
Mateusz Michalkiewicz, Jhony~K Pontes, Dominic Jack, Mahsa Baktashmotlagh, and
  Anders Eriksson.
\newblock Implicit surface representations as layers in neural networks.
\newblock In \emph{Proceedings of the IEEE/CVF International Conference on
  Computer Vision}, pages 4743--4752, 2019.

\bibitem[Modat et~al.(2014)Modat, Cash, Daga, Winston, Duncan, and
  Ourselin]{modat2014global}
Marc Modat, David~M Cash, Pankaj Daga, Gavin~P Winston, John~S Duncan, and
  S{\'e}bastien Ourselin.
\newblock Global image registration using a symmetric block-matching approach.
\newblock \emph{Journal of Medical Imaging}, 1\penalty0 (2):\penalty0 024003,
  2014.

\bibitem[Mok and Chung(2020)]{mok2020fast}
Tony~CW Mok and Albert Chung.
\newblock Fast symmetric diffeomorphic image registration with convolutional
  neural networks.
\newblock In \emph{Proceedings of the IEEE/CVF conference on computer vision
  and pattern recognition}, pages 4644--4653, 2020.

\bibitem[Muntoni and Cignoni(2021)]{pymeshlab}
Alessandro Muntoni and Paolo Cignoni.
\newblock {PyMeshLab}, January 2021.

\bibitem[Niemeyer et~al.(2019)Niemeyer, Mescheder, Oechsle, and
  Geiger]{niemeyer2019occupancy}
Michael Niemeyer, Lars Mescheder, Michael Oechsle, and Andreas Geiger.
\newblock Occupancy flow: 4d reconstruction by learning particle dynamics.
\newblock In \emph{Proceedings of the IEEE/CVF International Conference on
  Computer Vision}, pages 5379--5389, 2019.

\bibitem[Niemeyer et~al.(2020)Niemeyer, Mescheder, Oechsle, and
  Geiger]{niemeyer2020differentiable}
Michael Niemeyer, Lars Mescheder, Michael Oechsle, and Andreas Geiger.
\newblock Differentiable volumetric rendering: Learning implicit 3d
  representations without 3d supervision.
\newblock In \emph{Proceedings of the IEEE/CVF Conference on Computer Vision
  and Pattern Recognition}, pages 3504--3515, 2020.

\bibitem[Niu et~al.(2018)Niu, Li, and Xu]{niu2018im2struct}
Chengjie Niu, Jun Li, and Kai Xu.
\newblock Im2struct: Recovering 3d shape structure from a single rgb image.
\newblock In \emph{Proceedings of the IEEE conference on computer vision and
  pattern recognition}, pages 4521--4529, 2018.

\bibitem[Pan et~al.(2018)Pan, Li, Han, and Jia]{pan2018residual}
Junyi Pan, Jun Li, Xiaoguang Han, and Kui Jia.
\newblock Residual meshnet: Learning to deform meshes for single-view 3d
  reconstruction.
\newblock In \emph{2018 International Conference on 3D Vision (3DV)}, pages
  719--727. IEEE, 2018.

\bibitem[Pan et~al.(2019)Pan, Han, Chen, Tang, and Jia]{Pan_2019_ICCV}
Junyi Pan, Xiaoguang Han, Weikai Chen, Jiapeng Tang, and Kui Jia.
\newblock Deep mesh reconstruction from single rgb images via topology
  modification networks.
\newblock In \emph{Proceedings of the IEEE/CVF International Conference on
  Computer Vision (ICCV)}, October 2019.

\bibitem[Park et~al.(2019)Park, Florence, Straub, Newcombe, and
  Lovegrove]{park2019deepsdf}
Jeong~Joon Park, Peter Florence, Julian Straub, Richard Newcombe, and Steven
  Lovegrove.
\newblock Deepsdf: Learning continuous signed distance functions for shape
  representation.
\newblock In \emph{Proceedings of the IEEE/CVF Conference on Computer Vision
  and Pattern Recognition}, pages 165--174, 2019.

\bibitem[Paschali et~al.(2019)Paschali, Gasperini, Roy, Fang, and
  Navab]{paschali20193dq}
Magdalini Paschali, Stefano Gasperini, Abhijit~Guha Roy, Michael Y-S Fang, and
  Nassir Navab.
\newblock 3dq: Compact quantized neural networks for volumetric whole brain
  segmentation.
\newblock In \emph{International Conference on Medical Image Computing and
  Computer-Assisted Intervention}, pages 438--446. Springer, 2019.

\bibitem[Qi et~al.(2017)Qi, Su, Mo, and Guibas]{qi2017pointnet}
Charles~R Qi, Hao Su, Kaichun Mo, and Leonidas~J Guibas.
\newblock Pointnet: Deep learning on point sets for 3d classification and
  segmentation.
\newblock In \emph{Proceedings of the IEEE conference on computer vision and
  pattern recognition}, pages 652--660, 2017.

\bibitem[Ravi et~al.(2020)Ravi, Reizenstein, Novotny, Gordon, Lo, Johnson, and
  Gkioxari]{ravi2020pytorch3d}
Nikhila Ravi, Jeremy Reizenstein, David Novotny, Taylor Gordon, Wan-Yen Lo,
  Justin Johnson, and Georgia Gkioxari.
\newblock Accelerating 3d deep learning with pytorch3d.
\newblock \emph{arXiv:2007.08501}, 2020.

\bibitem[Rimol et~al.(2012)Rimol, Nesv{\aa}g, Hagler~Jr, Bergmann,
  Fennema-Notestine, Hartberg, Haukvik, Lange, Pung, Server,
  et~al.]{rimol2012cortical}
Lars~M Rimol, Ragnar Nesv{\aa}g, Don~J Hagler~Jr, {\O}rjan Bergmann, Christine
  Fennema-Notestine, Cecilie~B Hartberg, Unn~K Haukvik, Elisabeth Lange,
  Chris~J Pung, Andres Server, et~al.
\newblock Cortical volume, surface area, and thickness in schizophrenia and
  bipolar disorder.
\newblock \emph{Biological psychiatry}, 71\penalty0 (6):\penalty0 552--560,
  2012.

\bibitem[Rodriguez-Carranza et~al.(2008)Rodriguez-Carranza, Mukherjee,
  Vigneron, Barkovich, and Studholme]{rodriguez2008framework}
Claudia~E Rodriguez-Carranza, Pratik Mukherjee, D~Vigneron, J~Barkovich, and
  Colin Studholme.
\newblock A framework for in vivo quantification of regional brain folding in
  premature neonates.
\newblock \emph{Neuroimage}, 41\penalty0 (2):\penalty0 462--478, 2008.

\bibitem[Ronneberger et~al.(2015)Ronneberger, Fischer, and
  Brox]{ronneberger2015u}
Olaf Ronneberger, Philipp Fischer, and Thomas Brox.
\newblock U-net: Convolutional networks for biomedical image segmentation.
\newblock In \emph{International Conference on Medical image computing and
  computer-assisted intervention}, pages 234--241. Springer, 2015.

\bibitem[Roy et~al.(2019)Roy, Conjeti, Navab, Wachinger, Initiative,
  et~al.]{roy2019quicknat}
Abhijit~Guha Roy, Sailesh Conjeti, Nassir Navab, Christian Wachinger,
  Alzheimer's Disease~Neuroimaging Initiative, et~al.
\newblock Quicknat: A fully convolutional network for quick and accurate
  segmentation of neuroanatomy.
\newblock \emph{NeuroImage}, 186:\penalty0 713--727, 2019.

\bibitem[Santa~Cruz et~al.(2021)Santa~Cruz, Lebrat, Bourgeat, Fookes, Fripp,
  and Salvado]{cruz2021deepcsr}
Rodrigo Santa~Cruz, Leo Lebrat, Pierrick Bourgeat, Clinton Fookes, Jurgen
  Fripp, and Olivier Salvado.
\newblock Deepcsr: A 3d deep learning approach for cortical surface
  reconstruction.
\newblock In \emph{Proceedings of the IEEE/CVF Winter Conference on
  Applications of Computer Vision}, pages 806--815, 2021.

\bibitem[Schaer et~al.(2008)Schaer, Cuadra, Tamarit, Lazeyras, Eliez, and
  Thiran]{schaer2008surface}
Marie Schaer, Meritxell~Bach Cuadra, Lucas Tamarit, Fran{\c{c}}ois Lazeyras,
  Stephan Eliez, and Jean-Philippe Thiran.
\newblock A surface-based approach to quantify local cortical gyrification.
\newblock \emph{IEEE transactions on medical imaging}, 27\penalty0
  (2):\penalty0 161--170, 2008.

\bibitem[Segonne et~al.(2007)Segonne, Pacheco, and Fischl]{Segonne:TMI07}
Florent Segonne, Jenni Pacheco, and Bruce Fischl.
\newblock Geometrically accurate topology-correction of cortical surfaces using
  nonseparating loops.
\newblock \emph{IEEE Transactions on Medical Imaging}, 26\penalty0
  (4):\penalty0 518--529, 2007.

\bibitem[Shen et~al.(2019)Shen, Vialard, and Niethammer]{shen2019region}
Zhengyang Shen, Fran{\c{c}}ois-Xavier Vialard, and Marc Niethammer.
\newblock Region-specific diffeomorphic metric mapping.
\newblock \emph{arXiv preprint arXiv:1906.00139}, 2019.

\bibitem[Smith et~al.(2019)Smith, Fujimoto, Romero, and Meger]{smith19a}
Edward Smith, Scott Fujimoto, Adriana Romero, and David Meger.
\newblock {GEOM}etrics: Exploiting geometric structure for graph-encoded
  objects.
\newblock In Kamalika Chaudhuri and Ruslan Salakhutdinov, editors,
  \emph{Proceedings of the 36th International Conference on Machine Learning},
  volume~97 of \emph{Proceedings of Machine Learning Research}, pages
  5866--5876, Long Beach, California, USA, 09--15 Jun 2019. PMLR.

\bibitem[Su et~al.(2015)Su, Zeng, Wang, Lu, and Gu]{su2015shape}
Zhengyu Su, Wei Zeng, Yalin Wang, Zhong-Lin Lu, and Xianfeng Gu.
\newblock Shape classification using wasserstein distance for brain morphometry
  analysis.
\newblock In \emph{International Conference on Information Processing in
  Medical Imaging}, pages 411--423. Springer, 2015.

\bibitem[Taha and Hanbury(2015)]{taha2015metrics}
Abdel~Aziz Taha and Allan Hanbury.
\newblock Metrics for evaluating 3d medical image segmentation: analysis,
  selection, and tool.
\newblock \emph{BMC medical imaging}, 15\penalty0 (1):\penalty0 1--28, 2015.

\bibitem[Tatarchenko et~al.(2017)Tatarchenko, Dosovitskiy, and
  Brox]{tatarchenko2017octree}
Maxim Tatarchenko, Alexey Dosovitskiy, and Thomas Brox.
\newblock Octree generating networks: Efficient convolutional architectures for
  high-resolution 3d outputs.
\newblock In \emph{Proceedings of the IEEE International Conference on Computer
  Vision}, pages 2088--2096, 2017.

\bibitem[Trouv{\'e}(1995)]{trouve1995infinite}
Alain Trouv{\'e}.
\newblock An infinite dimensional group approach for physics based models in
  pattern recognition.
\newblock \emph{preprint}, 1995.

\bibitem[Vialard et~al.(2012)Vialard, Risser, Rueckert, and
  Cotter]{vialard2012diffeomorphic}
Fran{\c{c}}ois-Xavier Vialard, Laurent Risser, Daniel Rueckert, and Colin~J
  Cotter.
\newblock Diffeomorphic 3d image registration via geodesic shooting using an
  efficient adjoint calculation.
\newblock \emph{International Journal of Computer Vision}, 97\penalty0
  (2):\penalty0 229--241, 2012.

\bibitem[Wang et~al.(2018{\natexlab{a}})Wang, Zhang, Li, Fu, Liu, and
  Jiang]{wang2018pixel2mesh}
Nanyang Wang, Yinda Zhang, Zhuwen Li, Yanwei Fu, Wei Liu, and Yu-Gang Jiang.
\newblock Pixel2mesh: Generating 3d mesh models from single rgb images.
\newblock In \emph{Proceedings of the European Conference on Computer Vision
  (ECCV)}, pages 52--67, 2018{\natexlab{a}}.

\bibitem[Wang et~al.(2018{\natexlab{b}})Wang, Sun, Liu, and
  Tong]{wang2018adaptive}
Peng-Shuai Wang, Chun-Yu Sun, Yang Liu, and Xin Tong.
\newblock Adaptive o-cnn: a patch-based deep representation of 3d shapes.
\newblock \emph{ACM Transactions on Graphics (TOG)}, 37\penalty0 (6):\penalty0
  1--11, 2018{\natexlab{b}}.

\bibitem[Wickramasinghe et~al.(2020)Wickramasinghe, Remelli, Knott, and
  Fua]{wickramasinghe2020voxel2mesh}
Udaranga Wickramasinghe, Edoardo Remelli, Graham Knott, and Pascal Fua.
\newblock Voxel2mesh: 3d mesh model generation from volumetric data.
\newblock In \emph{International Conference on Medical Image Computing and
  Computer-Assisted Intervention}, pages 299--308. Springer, 2020.

\bibitem[Wu et~al.(2018)Wu, Zhang, Zhang, Zhang, Freeman, and
  Tenenbaum]{wu2018learning}
Jiajun Wu, Chengkai Zhang, Xiuming Zhang, Zhoutong Zhang, William~T Freeman,
  and Joshua~B Tenenbaum.
\newblock Learning shape priors for single-view 3d completion and
  reconstruction.
\newblock In \emph{Proceedings of the European Conference on Computer Vision
  (ECCV)}, pages 646--662, 2018.

\bibitem[Xu et~al.(2019)Xu, Wang, Ceylan, Mech, and Neumann]{Xu:NIPS19}
Qiangeng Xu, Weiyue Wang, Duygu Ceylan, Radomir Mech, and Ulrich Neumann.
\newblock Disn: Deep implicit surface network for high-quality single-view 3d
  reconstruction.
\newblock In \emph{Advances in Neural Information Processing Systems 32}, pages
  492--502, 2019.

\bibitem[Zhang et~al.(2020)Zhang, Liu, Zheng, and Shi]{zhang2020diffeomorphic}
Shuo Zhang, Peter~Xiaoping Liu, Minhua Zheng, and Wen Shi.
\newblock A diffeomorphic unsupervised method for deformable soft tissue image
  registration.
\newblock \emph{Computers in biology and medicine}, 120:\penalty0 103708, 2020.

\bibitem[Zubi{\'c} and Li{\`o}(2021)]{zubic2021effective}
Nikola Zubi{\'c} and Pietro Li{\`o}.
\newblock An effective loss function for generating 3d models from single 2d
  image without rendering.
\newblock \emph{arXiv preprint arXiv:2103.03390}, 2021.

\end{thebibliography}

\end{document}